\documentclass[runningheads]{llncs}

\usepackage{eccv}

\usepackage{eccvabbrv}

\usepackage{graphicx}
\usepackage{booktabs}

\usepackage[accsupp]{axessibility}  

\usepackage{hyperref}

\usepackage{orcidlink}

\usepackage{multirow}
\usepackage{xcolor}
\usepackage{colortbl}
\usepackage{adjustbox}
\usepackage{pifont}
\usepackage{makecell}
\usepackage{array}
\usepackage{tikz}
\usepackage{amssymb}
\usepackage{mathabx}
\usepackage{fontawesome}

\definecolor{my_green}{RGB}{127,255,0}
\definecolor{my_red}{RGB}{204, 0, 0}
\definecolor{my_gray}{RGB}{128, 128, 128}
\renewcommand{\checkmark}{\textcolor{my_green}{\ding{51}}} 

\newcommand{\checkmarkgray}{\textcolor{my_gray}{\ding{51}}} 

\newcommand{\crossmark}{\textcolor{my_red}{\ding{55}}} 

\begin{document}

\title{Beyond Description: Cognitively Benchmarking Fine-Grained Action for Embodied Agents} 

\titlerunning{CFG-Bench}

\author{Dayong Liu$^1$$^,$$^2$$^{\star\ddagger}$
\and Chao Xu$^2$$^,$$^3$$^{\star}$
\and Weihong Chen$^2$$^,$$^3$
\and Suyu Zhang$^2$$^,$$^3$
\and \\Juncheng Wang$^4$ 
\and Jiankang Deng$^5$ 
\and Baigui Sun$^2$$^,$$^3$$^{\dag}$
\and Yang Liu$^2$$^,$$^3$$^{\dag}$\\
}

\authorrunning{D.~Liu et al.}

\institute{Zhejiang University \and
IROOTECH TECHNOLOGY \and
Wolf 1069 b Lab, Sany Group \and
The Hong Kong Polytechnic University \and
Imperial College London\\[0.6em]
\email{liu\_dayong@zju.edu.cn, \{baigui.sun, yang.liu1\}@irootech.com}
}

\maketitle
\begingroup\renewcommand\thefootnote{}\footnotetext{$^{\ddagger}$~Work done during an internship at IROOTECH TECHNOLOGY. \\$^{\star}$~Equal contribution.\quad
$^{\dagger}$~Corresponding authors. }\addtocounter{footnote}{-1}\endgroup

\begin{abstract}
Multimodal Large Language Models (MLLMs) show promising results as decision-making engines for embodied agents operating in complex, physical environments. However, existing benchmarks often prioritize high-level planning or spatial reasoning, leaving the fine-grained action intelligence required for embodied physical interaction underexplored. To address this gap, we introduce CFG-Bench, a new benchmark designed to systematically evaluate this crucial capability. CFG-Bench consists of 1,368 curated videos paired with 19,562 question-answer pairs spanning three evaluation paradigms targeting four cognitive abilities: 1) Physical Interaction, 2) Temporal-Causal Relation, 3) Intentional Understanding, and 4) Evaluative Judgment. Together, these dimensions provide a systematic framework for assessing a model's ability to translate visual observations into actionable knowledge, moving beyond mere surface-level recognition. Our comprehensive evaluation on CFG-Bench reveals that leading MLLMs struggle to produce detailed instructions for physical interactions and exhibit profound limitations in the higher-order reasoning of intention and evaluation. Moreover, supervised fine-tuning (SFT) on our data demonstrates that teaching an MLLMs to articulate fine-grained actions directly translates to significant performance gains on established embodied benchmarks. Our analysis highlights these limitations and offers insights for developing more capable and grounded embodied agents. Project page: \url{https://cfg-bench.github.io/}
\end{abstract}    
\section{Introduction}
\label{sec:intro}
\begin{figure}[t!]
	\centering
    \includegraphics[width=0.98\textwidth]{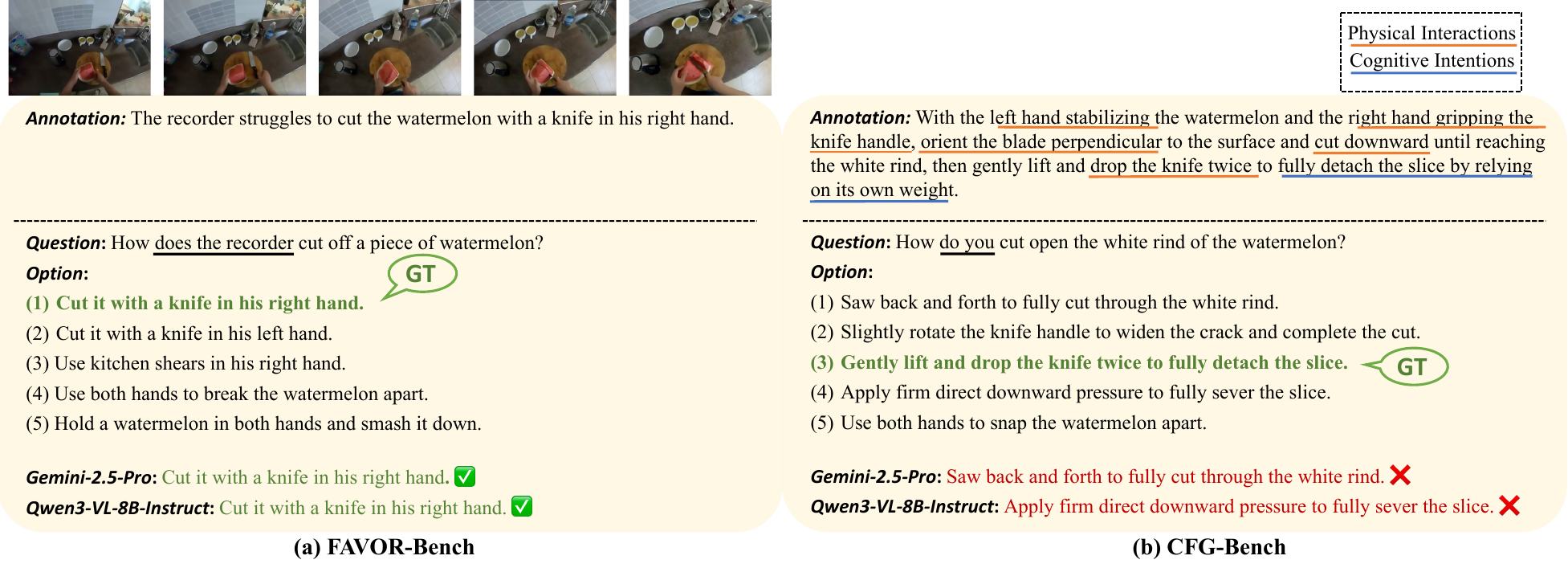}
	\caption{Illustration of CFG-Bench's focus on embodied intelligence over descriptive accuracy. The part (a) shows how FAVOR-Bench annotates and questions from a third-person perspective for motion-level actions, a task which current MLLMs can often solve. In contrast, the part (b) demonstrates CFG-Bench's fine-grained annotation and first-person scenario questions, which probes for the actionable physical and intentional details necessary for embodied agents. Current MLLMs struggle to master the crucial fine-grained details required for physical interaction.}
	\label{fig:teaser}
\end{figure}
The remarkable capabilities of Multimodal Large Language Models (MLLMs)~\cite{Videollama, Qwen2.5-VL, Internvl3} in understanding and reasoning across vision and language have driven their active exploration as the decision-making engine for embodied agents~\cite{liang2022code, singh2022progprompt, song2023llm, han2025multimodal, jin2024reasoning, shao2025large}. Operating within interactive environments, these agents are required to interpret multimodal observations to inform their planing and guide their subsequent actions. To systematically evaluate these capabilities, a variety of benchmarks have emerged~\cite{Embodiedeval, yang2025thinking, dang2025ecbench, Embodiedbench, qi2025bear}, with a significant focus on areas like egocentric perception~\cite{Egothink}, visual grounding~\cite{zhang2024task}, and spatial reasoning~\cite{Spatialvlm}.

\begin{table*}[t!]
\centering
\caption{Comparison with other benchmarks. CFG-Bench introduces a four-tiered cognitive framework for embodied fine-grained intelligence, distinguishing it from existing action benchmarks focused on third-person coarse description and other embodied benchmarks that prioritize spatial reasoning and high-level planing over the fine-grained physical action. It also provides a more comprehensive evaluation protocol by integrating three evaluation paradigms of QAs. \checkmarkgray indicates partial coverage.}
\resizebox{1\linewidth}{!}
{
\renewcommand{\arraystretch}{1.0} 
\begin{tabular}{lcccccccccc}
\toprule
{\multirow{2}{*}{\textbf{Benchmarks}}} & {\multirow{2}{*}{\textbf{Embodied}}} & \multicolumn{4}{c}{\textbf{Fine-Grained Action}} & {\multirow{2}{*}{\textbf{Data Scale}}} & \multicolumn{4}{c}{\textbf{QA Pairs}}  \\
\cmidrule{3-6} \cmidrule{8-11}
 & & \textbf{Interaction} & \textbf{Temporal-Causal} & \textbf{Intention} & \textbf{Evaluation} & & \textbf{Number} & \textbf{Closed-Ended} & \textbf{Open-Ended} & \textbf{Counterfactuals} \\

\midrule
MVBench~\cite{Mvbench} & \crossmark & \checkmarkgray & \checkmark & \crossmark & \crossmark & 4,000 & 4,000 & \checkmark & \crossmark  & \checkmark\\
VideoMME~\cite{Videomme} & \crossmark & \checkmarkgray & \checkmark & \crossmark & \crossmark & 2,525 & 2,525 & \checkmark & \crossmark & \crossmark \\
EgoSchema~\cite{Egoschema} & \crossmark & \checkmarkgray & \checkmark & \crossmark & \crossmark & 5,031 & 5,031 & \checkmark & \crossmark & \crossmark \\
EgoTaskQA~\cite{Egotaskqa} & \crossmark & \checkmarkgray & \checkmark & \checkmark & \crossmark & 2,315 & 40,322 & \checkmark & \checkmark  & \checkmark \\
Motion-Bench~\cite{Motionbench} & \crossmark & \checkmarkgray & \checkmarkgray & \crossmark & \crossmark & 5,385 & 8,052 & \checkmark & \crossmark & \crossmark \\
FAVOR-Bench~\cite{Favorbench} & \crossmark & \checkmarkgray & \checkmarkgray & \crossmark & \crossmark & 1,776 & 8,184 & \checkmark & \checkmark & \crossmark \\
VSI-Bench~\cite{yang2025thinking} & \checkmark & - & - & - & - & 288 & 5,000 & \checkmark & \checkmark & \crossmark\\
ECBench~\cite{dang2025ecbench} & \checkmark & - & - & - & - & 386 & 4,324 & \checkmark & \checkmark & \checkmark\\
EMBODIEDEVAL~\cite{Embodiedeval} & \checkmark & - & - & - & - & 328 & 1,533 & \checkmark & \crossmark & \checkmark\\
\midrule
\textbf{CFG-Bench} & \checkmark & \checkmark & \checkmark & \checkmark & \checkmark & 1,368 & 19,562 & \checkmark & \checkmark & \checkmark \\
\bottomrule
\end{tabular}
}
\label{tab:comparison}
\end{table*}

However, while existing benchmarks effectively assess an agent's visual-spatial intelligence and strategic planning, they largely overlook the most challenging dimension of interaction: \textit{fine-grained action intelligence}, which refers to comprehending the details of how an action is executed, both \textit{physically} and \textit{cognitively}. Probing these fine-grained details is crucial for pushing embodied imitation learning~\cite{cui2022play, wang2023mimicplay} beyond rigid trajectory replication, as it tests an agent's ability to perform executable physical actions as well as high-level planning and reasoning required to generalize successfully to various scenarios.

Concurrently, benchmarks within the action understanding domain have begun to recognize the importance of fine-grained detail. Pioneering work like FAVOR-Bench~\cite{Favorbench}, has moved beyond event-level labels to probe motion-level perception and the precise temporal sequence of actions. However, from an embodied perspective, their approach is limited in two ways. 
(1) The definition of fine-grained is often not granular enough for physical execution. As shown in Fig.~\ref{fig:teaser}, the current annotation simply states ``cut the watermelon with a knife in right hand,'' yet fails to capture granular execution sequences, e.g., ``stabilize the watermelon with the left hand, grip the knife handle with the right hand and then cut down vertically with the blade down.''
(2) It is oriented towards objective video description from a third-person perspective, does not encompass the deep cognitive reasoning underlying surface-level action, such as ``gently lift and drop knife twice'' is to fully detach the slice by relying on its own weight.

To systematically address these challenges, we first synthesize a hierarchical framework grounded in Norman’s Action Cycle~\cite{norman2013design} that deconstructs fine-grained action intelligence into four hierarchical tiers: (1) \textit{Physical Interaction}, detailing how an action is physically executed (aligned with  Normman's Specifying and Performing); (2) \textit{Temporal-Causal Relation}, reasoning how actions connect through time and causality (Normman's Sequencing and Planning); (3) \textit{Intentional Understanding}, inferring why an action is performed (Norman's Goal Formation); and (4) \textit{Evaluative Judgment}, evaluating how well an action was done and how to improve it if necessary (Norman's Comparing Outcome with Goal). This structure helps an embodied agent internalize the closed loop among formulation, planning, execution, and feedback from demonstrations, transforming surface-level recognition into cognitive executable knowledge.

Building on this four-tiered framework, we introduce CFG-Bench, \textbf{c}ognitively benchmarking \textbf{f}ine-\textbf{g}rained action for embodied agents. Specifically, 
CFG-Bench includes a curated dataset of 1,368 videos, covering ego- and exo-centric daily-life records, hand-object interactions, and more complex outdoor activities. 
We employ close-ended multiple-choice question-answering (QA) for more objective Physical Interaction and Temporal-Causal Relation tiers, while adopting an open-ended format for the higher-order Intentional Understanding and Evaluative Judgment tiers, which is crucial to prevent models from reasoning backward from the textual options, forcing them to ground their inference in the visual evidence. 
Notably, we integrate open-ended counterfactual questions across all four tiers to directly combat model hallucination and probe a deeper action comprehending. 
All QA pairs undergo a rigorous human-in-the-loop verification process to eliminate overly simple or erroneous questions. 
Overall, CFG-Bench comprises 19,562 QA pairs distributed across 11 distinct tasks within the 4 tiers. 
For open-ended tasks, we use the standard GPT-assisted evaluation~\cite{sun2024aligning, zhou2024mlvu}, but apply a gating mechanism for parts of counterfactuals: a response is scored only if the model first identifies the question's false premise. Furthermore, since both annotation and evaluation involve LLMs, we include Inter-Annotator agreement and Human-LLM agreement to validate the reliability of LLM-assisted ground truth annotations and GPT-assisted evaluation, respectively. 

Finally, we conduct a comprehensive zero-shot evaluation of leading proprietary and open-source MLLMs on CFG-Bench. Furthermore, by fine-tuning Qwen2.5-VL~\cite{Qwen2.5-VL} on our data, we demonstrate how a model that has learned to articulate fine-grained actions with physical and intentional details shows substantial performance gains on established benchmarks for embodied manipulation and planning~\cite{Embodiedbench}.

Our contribution can be concluded from three aspects:
\begin{itemize}
    \item We define fine-grained action intelligence and propose a novel four-tiered framework to systematically deconstruct and evaluate it beyond mere objective description.
    \item We construct CFG-Bench, a new benchmark featuring a hybrid QA design and innovative open-ended counterfactual challenges to rigorously test the physical and cognitive grounding of MLLMs on fine-grained actions.
    \item We provide a comprehensive analysis revealing the limitations of current MLLMs and validate that training on CFG-Bench significantly enhances planning and manipulation capabilities for embodied physical execution.
\end{itemize}
\section{Related Works}
\label{sec:related}
\subsection{Benchmarks for Embodied Agent}
As MLLMs are increasingly explored as the ``brains'' for robots~\cite{Robobrain, Robobrain2.0, mu2023embodiedgpt, openeqa, pcabench}, several benchmarks have emerged to evaluate their ability to perceive and reason about the physical world. For instance, VSI-Bench~\cite{yang2025thinking} is designed to assess the visual-spatial intelligence of MLLMs, probing their understanding of object positions, orientations, and geometric relations. Similarly, ECBench~\cite{dang2025ecbench} is dedicated to evaluating embodied cognitive abilities within both static and dynamic environments. These benchmarks are key to an agent’s spatial perception and mapping baseline.

A second category, offline benchmarks focus on what high-level action/plan, e.g., Ego-PlanBench~\cite{egoplan} for ego symbolic planning; EgoExoBench~\cite{egoexobench} for ego-exo alignment; RoboBench~\cite{robobench} and BEAR~\cite{qi2025bear} for task- and atomic-level embodied capability diagnostics. 
The recent online benchmarks like EMBODIEDEVAL~\cite{Embodiedeval} and EMBODIEDBENCH~\cite{Embodiedbench} introduce a more comprehensive suite of diverse tasks and scenes to evaluate whether the task succeeds. They abstract away the complexity of physical interaction. The former focuses on the strategic what to do next by reducing actions to high-level, symbolic commands (e.g., pick up). The latter's manipulation tasks require the MLLM to act as a direct, end-to-end policy that outputs low-level commands, such as a 7-DoF vector. We argue that our approach, which focuses on generating cognitive knowledge about how and why an action should be performed, is deeply complementary to these works. The experiments prove that an MLLM fine-tuned on our data achieves significant performance gains on the high-level planning and low-level control tasks in Sec.~\ref{sec:4.4}.

\subsection{Benchmarks for Action Understanding}
The evolution of action understanding benchmarks reveals a progression towards greater detail, yet a critical gap persists between understanding for descriptive analysis versus for embodied execution. Early comprehensive benchmarks like MVBench~\cite{Mvbench} and Video-MME~\cite{Videomme} include subsets for action, but these are situated within a broader evaluation of general video understanding. A more dedicated line of work, exemplified by ActivityNet-QA~\cite{Activitynet-qa}, concentrates on event-level granularity, requiring models to identify the main activity or sequence of major events, but still lacks consideration for more fine-grained actions.

To address this, pioneering works like EgoTaskQA~\cite{Egotaskqa}, MotionBench~\cite{Motionbench}, MotionSight~\cite{du2025motionsight}, and FAVOR-Bench~\cite{Favorbench} move beyond event labels and attend to motion-level perception and precise temporal sequences. However, their definition of fine-grained is still primarily oriented towards achieving the fidelity of a third-person observer, targeting high-accuracy video description. They do not systematically evaluate the deep procedural knowledge required for an embodied agent to physically replicate an action, like the specific contact details and dynamics, and the causal, intentional, and evaluative reasoning that underpins skillful interaction. This deeper layer of fine-grained action intelligence, which separates merely observing an action from truly understanding how and why to perform it, remains the unaddressed frontier.
\section{CFG-Bench}
\label{sec:bench}
\begin{figure*}[t!]
	\centering
    \includegraphics[width=0.98\textwidth]{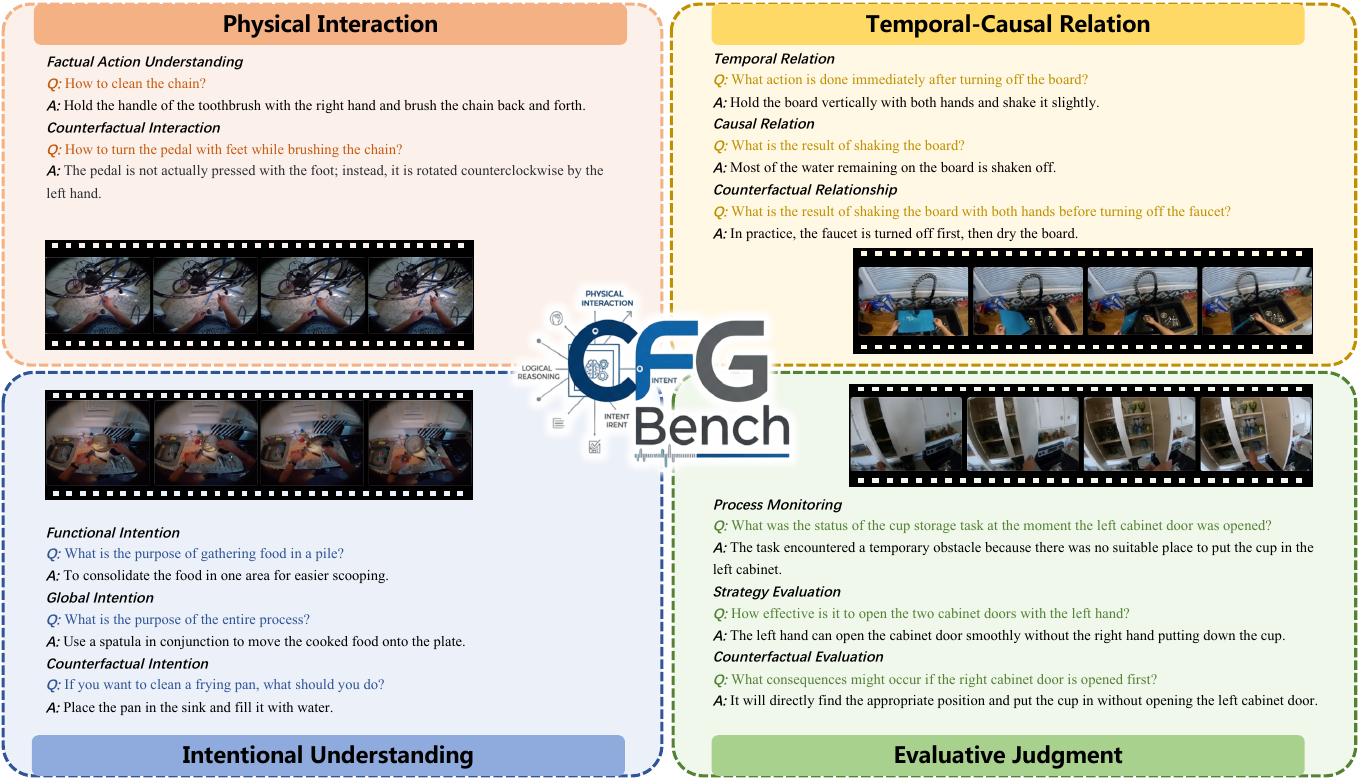}
	\caption{Task demonstration of CFG-Bench. Notably, all QA pairs, including those above, are slightly simplified for clarity and brevity.}
	\label{fig:demo}
\end{figure*}

An embodied agent's ability to translate multimodal inputs into successful physical interactions hinges on its fine-grained action intelligence. Encompassing the physical how and cognitive why of an action, this capability serves as the essential bridge between abstract planning and low-level physical control. To this end, we develop CFG-Bench to systematically measure this intelligence. 
First, we deconstruct the fine-grained action into 4 embodied cognitive tiers (Sec.~\ref{sec:3.1}). We then detail the dataset construction process, including video curation and QA generation (Sec.~\ref{sec:3.2}). 
Finally, we describe the evaluation protocols for QAs (Sec.~\ref{sec:3.3}).
Fig.~\ref{fig:demo} is an overview of CFG-Bench tasks.

\subsection{Taxonomy and Task Design}
\label{sec:3.1}
\noindent{\textbf{Physical Interaction}}
is the ability to comprehend the physical mechanics of an action can move beyond simple event recognition to a detailed, factual understanding of how that action is physically performed. This granular perception is the foundation for an agent's ability to imitate, interact with, and learn from the physical world.

Concretely, \textit{Factual Action Understanding} (FAU) is designed to measure the model's ability to extract the actionable and fine-grained details of physical interaction from visual inputs. Through multi-choice questions (MCQ), this task cover a full spectrum of the specific agent performing the action (e.g., left hand or right hand), the objects and tools being manipulated (e.g., cup or bottle), the specific parts of an object being interacted with (e.g., handle or rim), the precise operation type (e.g., gripping or holding), and the dynamic properties of the motion (e.g., forward quickly or backward slowly). 

We further introduce \textit{Counterfactual Interaction} (CIA) to test the model's robustness and its ability to ground responses firmly in visual evidence through open-ended QA. This task confronts the model with a question that contains a false premise. A successful response requires the model to identify this false premise as inconsistent with the video and provide a correction based on what actually occurred. This method directly penalizes the common failure mode of acquiescence, where models blindly accept and hallucinate based on incorrect information. 

\noindent{\textbf{Temporal-Causal Relation}} elevates from isolated actions to the logical structure that connects them, which indicates the model's ability to reason about how actions relate to each other in time and through cause-and-effect. This capability is a fundamental factor for an agent to plan, anticipate outcomes, and comprehend long-horizontal tasks.

This is specifically evaluated
by \textit{Temporal Relation} (TR) and \textit{Causal Relation} (CR). For the former, TR is designed to ensure a comprehensive assessment of the model's understanding of event timelines. We probe this capability through questions targeting sequential ordering (e.g., ``What occurred after action X?''), requiring holistic sequence verification (e.g., ``Which option correctly lists the event order?''), and testing for concurrent action identification (e.g., ``What was the left hand doing while the right hand performed the task?''). For the latter, CR focuses on the direct consequences of an action, asking the model to identify the immediate outcome resulting from a specific event. This tests its ability to reason about the consequences of its own potential actions. In addition, similar to CIA, we introduce the \textit{Counterfactual Relationship} (CRS) to ensure that the causal reasoning of the model is firmly grounded in the specific visual cues.

\noindent{\textbf{Intentional Understanding}} marks a significant shift from the how of an action to the why, which representing the model is capable of inferring the underlying goals and motivations that drive physical behavior, serving as a critical ability for an agent to formulate its sub-goals. To assess thorough reasoning and prevent models from simply selecting a plausible answer from a list, all tasks within this tier are exclusively formulated as open-ended questions.

We begin with \textit{Functional Intention} (FI), a task that focuses on the immediate purpose behind a specific action or its particular manner of execution. FI includes two question types, ``why performs a single atomic action'' or ``why adopt this specific physical technique''. In contrast to this granular focus, the \textit{Global Intention} (GI) task requires the model analyze a sequence of distinct and scattered actions to deduce the overall goal that connects. This challenge the model to understand how individual sub-tasks contribute to a high-level plan. Finally, the \textit{Counterfactual Intention} (CIT) task provides the most rigorous test of the model's grasp of how intention drives behavior. It presents the model with a hypothetical change to the agent's overall goal and ask it to predict the resulting changes in actions. 

\noindent{\textbf{Evaluative Judgment}} shifts the focus from comprehension to evaluation and is defined as the agent's ability to make qualitative assessments of an action's execution and outcome. This is a critical capability, as it is essential for both learning from observation by distinguishing effective from hazardous techniques, and for achieving resilience by identifying failures and re-planning accordingly. Since this reasoning requires justification, all tasks are open-ended.

Specifically, the first task is \textit{Process Monitoring} (PM), which requires the model analyze the ongoing action to determine if the process is proceeding as expected, or if it has encountered a problem, such as being temporarily stuck, having failed, or being unexpectedly interrupted. Next, the \textit{Strategy Evaluation} (SE) task requires the model to adopt the role of a critic, judging the quality of an action's execution based on criteria such as rationality, professionalism, and efficiency. Thus, it favors actions that are more likely to succeed in the specific scene. Finally, similar to the above preceding tiers, the \textit{Counterfactual Evaluation} (CEU) task confronts the model with a hypothetical change on a specific physical interaction or event within the process and asks it to evaluate how that change would have impacted the quality of the action or its outcome.

\subsection{Dataset Generation}
\label{sec:3.2}
\noindent\textbf{Video Source.}
Driven by the need to support evaluation across four cognitive tiers, we construct a final corpus from five datasets, each selected for its specific contributions.

\begin{figure}[t!]
	\centering
    \includegraphics[width=0.98\textwidth]{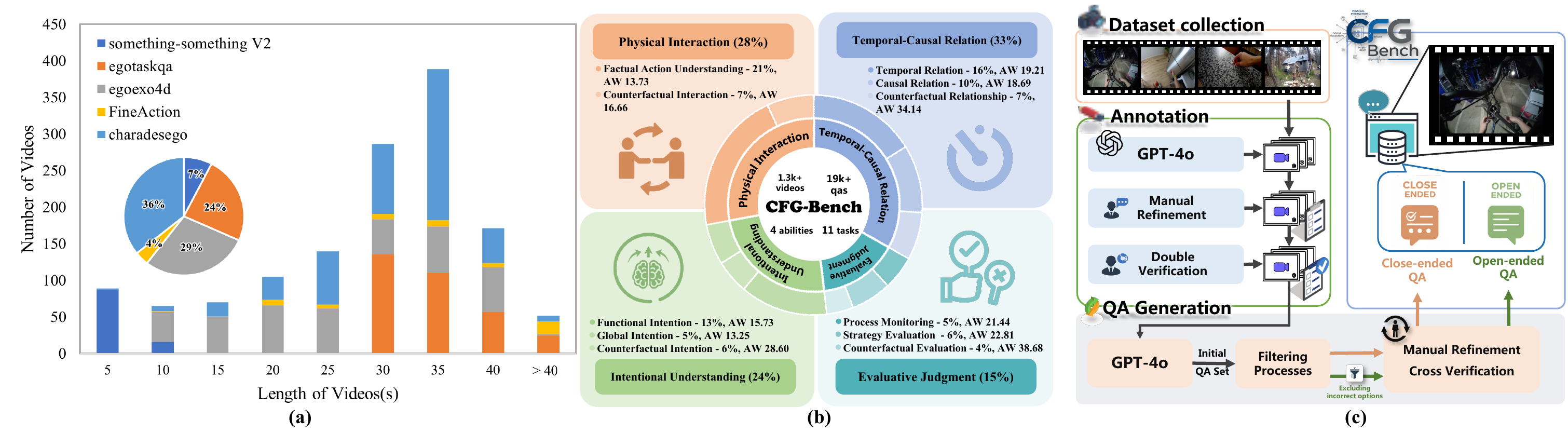}
	\caption{(a) Distribution and video length statistics of five datasets. Video-length min/mean/max = 1.83/26.3/107.8s (b) The distribution of tasks across four tiers. AW means average words of questions. (c) Pipeline of dataset generation. Both annotation and QA generation are human-AI collaborative workflow. Open-ended and closed-ended questions share the same pipeline at the early stage.}
	\label{fig:data_pipeline_statistic}
\end{figure}

We prioritize the first-person perspective, which is essential for embodied intelligence, selecting 329 videos from EgoTaskQA~\cite{Egotaskqa}, 191 from Charades-Ego~\cite{Charades}, and 394 from EgoExo4d~\cite{Ego-exo4d}. These datasets are rich in daily, goal-oriented tasks, offering clear examples of physical interaction (Tier 1), distinct temporal workflows and intentions (Tiers 2 \& 3), and scenarios for evaluation (Tier 4).
To complement this view, 298 third-person videos from Charades-Ego~\cite{Charades} provide an objective perspective ideal for analyzing causal-temporal relation and evaluating strategy (Tiers 2 \& 4). Besides, 104 videos from Something-Something-V2~\cite{ssv2} are included for their focus on fine-grained hand-object interactions, a key aspect of embodied manipulation, and 52 from FineAction~\cite{Fineaction} for its complex outdoor action diversity. Fig.~\ref{fig:data_pipeline_statistic} (a) shows the distribution of dataset categories and the statistics of video lengths.

\noindent\textbf{Manual Annotation.} To generate high-quality ground truth for our benchmark, we conduct a meticulous, month-long annotation process. As shown in Fig.~\ref{fig:data_pipeline_statistic} (c), our team of ten is organized into a two-stage pipeline that begin with draft annotations from GPT-4o~\cite{GPT-4o}. In the first stage, eight expert annotators refine these initial drafts, after which two expert reviewers perform a final verification. The entire process is guided by our four-tiered cognitive framework, with specific annotation instructions tailored to the unique characteristics of each source dataset. The annotation guideline is shown in Appendix~\ref{sec:C4}.

\noindent\textbf{Close-Ended QA.}
With the curated videos and its detailed annotations, we proceed to generate the QA pairs. For FAU, TR, and CR in Tiers 1 and 2, we develop a rigorous multi-stage pipeline to generate close-ended QA pairs.

As shown in Fig.~\ref{fig:data_pipeline_statistic} (c), the process begins with an automated generation phase where we design distinct prompt templates for each sub-task. Using GPT-4o~\cite{GPT-4o} (prompts are shown in Appendix~\ref{sec:C2}), we generate an initial set of QA pairs with two principles: maximize question diversity across the videos and craft challenging distractors that are plausible but unique. We obtain 19,372 multiple-choice QAs pairs in this stage.

Then, an automated filtering stage designed to eliminate overly simple questions. Following FAVOR-Bench~\cite{Favorbench}, we adopt the blind and single-frame filtering techniques, removing questions that could be answered with common sense alone or from a single static image. To further refine the set, we provide Qwen3-Max~\cite{qwen3} with the QAs and the corresponding video caption, which serves as a proxy for the ground truth. The model is tasked with flagging any QA pairs it deemed ambiguous or impossible to choose from, thereby efficiently identifying potentially flawed questions for further targeted human review.
After this process, approximately 35\% of the initially QA pairs are discarded.

The final stage is a manual verification. This process is conducted by six trained annotators. To guarantee the video-relevant of each question and the correctness and uniqueness of its answer, the team perform multiple rounds of validation. All QA pairs are cross-checked to resolve ambiguous references and reduce subjectivity. This human-in-the-loop process results in 9,165 QAs.

\noindent\textbf{Open-Ended QA.}
For the inferential tasks within our benchmark, i.e., the counterfactuals in Tiers 1 and 2, and all questions in Tiers 3 and 4, we develop a human-centric crafting process to generate high-quality open-ended QA pairs. As shown in Fig.~\ref{fig:data_pipeline_statistic} (c), our methodology begins by first generating a dedicated set of multiple-choice questions, obtaining 10,397 QAs. We then only maintain the question and its corresponding correct answer to serve as a factually-grounded seed for each open-ended query. This seed set is then manually refined by our ten annotators, who would either rephrase the question to be more exploratory and elaborate on the answer, or, if the initial seed is inadequate, craft a completely new QA pair from scratch. This process ensures every question is rooted in a verifiable evidence and paired with a reference answer containing the rich detail. All final pairs are validated through expert cross-checking either. 

\noindent\textbf{Statistics.}
CFG-Bench comprises 19,562 QA pairs, evenly split between 9,165 close-ended and 10,397 open-ended questions distributed across our 11 tasks. As detailed in Fig.~\ref{fig:data_pipeline_statistic} (b), the average words (AW) of counterfactual questions is significantly longer, as these queries require more context to elicit reasoned responses. The figure also highlights a notable variance in the number of questions per task, since not every video naturally suitable for all 11 task types. 

\subsection{Evaluation}
\label{sec:3.3}
We evaluate MLLM performance on CFG-Bench using a hybrid methodology. Close-ended tasks are measured by standard accuracy. For open-ended tasks, we follow~\cite{Favorbench} and use a GPT-assisted protocol to score responses on two dimensions: Correctness, which measures factual accuracy against the video, and Detailedness, which assesses descriptive richness, with the final score being their mean. DeepSeek-R1~\cite{guo2025deepseek} is employed in GPT-assisted evaluation and the prompt is attached in Appendix~\ref{sec:C3}. Notably, a strict gating mechanism applies to the counterfactual tasks in Tiers 1 and 2: a response is only eligible for scoring if it explicitly rejects or implicitly corrects the question's false premise. Failure to do so results in an automatic score of zero for both dimensions. 
This gating mechanism requires embodied agents to visually identify false premises before describing grounded reality, thus preventing blind compliance.

\section{Experiments}
\label{sec:experiments}
\subsection{Experimental Setup}
We benchmark a suite of leading MLLMs on CFG-Bench using their official implementations or APIs. Our evaluation includes proprietary models (Gemini-3-Pro~\cite{gemini3-pro}, Gemini-2.5-Pro~\cite{Gemini-2.5}, GPT-5~\cite{GPT-5}) and a range of open-source counterparts (Video-LLaMA3~\cite{Videollama}, InternVL3~\cite{Internvl3}, Gemma-3~\cite{Gemma3}, Qwen2.5-VL~\cite{Qwen2.5-VL}, Qwen3-VL~\cite{qwen3}). To specifically assess performance from an embodied perspective, we also include two specialized foundation models, RoboBrain2.0~\cite{Robobrain2.0} and Cosmos~\cite{cosmos}. To ensure a fair comparison, video frames for each model are sampled according to their officially recommended frame-rate-based strategies.
\renewcommand{\arraystretch}{1.2}
\begin{table*}[t]
\caption{Comprehensive results of leading proprietary and open-source MLLMs on CFG-Bench. It presents the performance of each task within every cognitive tier, along with the average score for open-ended questions $\textbf{Avg}_o$ and the average accuracy for close-ended questions $\textbf{Avg}_c$. Random selection and human performance are also included for comparison. The highest and suboptimal results are \textbf{bolded} and \underline{underlined}.} 
\small
\resizebox{\linewidth}{!}{
\begin{tabular}{lcccccccccccccc}
\toprule
\specialrule{0em}{0.3pt}{0.3pt}
\multirow{2}{*}{\textbf{Methods}}  & \multirow{2}{*}{\textbf{Input}}  & \multirow{2}{*}{\textbf{$Avg_{c}$}} & \multirow{2}{*}{\textbf{$Avg_{o}$}} & \multicolumn{2}{c}{\textbf{Interaction}} & \multicolumn{3}{c}{\textbf{Temporal-Causal}}  & \multicolumn{3}{c}{\textbf{Intention}} & \multicolumn{3}{c}{\textbf{Evaluation}}\\ 
\specialrule{0em}{0.3pt}{0.3pt}
\cmidrule(r){5-6} \cmidrule(r){7-9} \cmidrule(r){10-12} \cmidrule(r){13-15}
\specialrule{0em}{0.1pt}{0.1pt}
&~&~&~&FAU & CIA & TR & CR & CRS & FI & GI
& CIT  & PM & SE & CEU \\   
\specialrule{0em}{0.3pt}{0.3pt}
\hline
Human & -- &95.85  & 9.05 &93.48 &8.76 &95.97 &98.10  &9.02  &8.83  &9.51  &9.11  &8.92  &8.63 & 9.62    \\
Random & -- & -- & -- &20 &-- &20 &20 &--  &--  &--  &--  &--  &-- &--    \\

\hline
 \rowcolor[HTML]{E3F8F8}\multicolumn{15}{l}{\textcolor{gray}{{\textit{\textbf{Proprietary MLLMs}}}}}\\
Gemini-2.5-Pro~\cite{Gemini-2.5} & 4 fps & \underline{59.52} & \underline{5.40} & 58.81 & 4.58 & 54.81 & 64.95 & 4.55 & 5.18 & 6.15 & 5.31 & 5.72 & 5.19 & 6.49\\
Gemini-3-Pro~\cite{gemini3-pro} & 4 fps & \textbf{65.60} & \textbf{5.67} & 63.64 & 4.68 & 60.82 & 72.35 & 5.39 & 5.29 & 6.48 & 6.31 & 5.57 & 5.32 & 6.32\\
GPT-5~\cite{GPT-5} & 1 fps & 55.44 & 4.13 & 54.71 & 2.61 & 46.53 & 65.09 & 3.22 & 4.98 & 4.64 & 3.16 & 4.55 & 4.82 & 5.05\\
          
\hline
        
 \rowcolor[HTML]{FFF5F5}\multicolumn{15}{l}{\textcolor{gray}{{\textit{\textbf{Open-source MLLMs}}}}} \\ 

Video-LLaMA3-2B~\cite{Videollama} & 4 fps & 43.23 & 3.78 & 51.21 & 2.14 & 34.64 & 43.83 & 2.80 & 4.77 & 4.82 & 3.24 & 4.19 & 4.07 & 4.21\\
Video-LLaMA3-7B~\cite{Videollama} & 4 fps & 51.27 & 4.13  &51.33	&2.39	&50.17	&52.31	&3.29	&5.03	&4.96	&3.44	&4.68	&4.21	&5.05\\
InternVL3-2B~\cite{Internvl3} & 4 fps & 41.02 & 4.03 &53.46	&2.61	&33.90	&35.70	&3.13	&4.05	&4.73	&3.12	&4.62	&4.83	&5.15\\
InternVL3-8B~\cite{Internvl3} & 4 fps & 43.54 &4.13 & 56.26	&2.63	&36.89	&37.48	&3.15	&5.03	&4.67	&3.13	&4.56	&4.85	&5.02\\
InternVL3-78B~\cite{Internvl3} & 4 fps & 52.82	&\underline{4.92}	&58.31	&3.95	&49.42	&50.74	&3.68	&5.41	&5.04	&4.80	&4.96	&5.27	&6.21\\
Gemma-3-4B~\cite{Gemma3} & 4 fps & 48.49	&3.73	&44.43	&2.46	&46.44	&54.59	&2.86	&4.61	&2.44	&3.24	&4.22	&4.51	&5.46\\
Gemma-3-12B~\cite{Gemma3} & 4 fps &50.05	&4.07	&47.87	&2.53	&47.94	&54.33	&2.95	&5.20	&2.51	&4.36	&4.60	&4.69	&5.74\\
Gemma-3-27B~\cite{Gemma3} & 4 fps & \underline{52.96}	& 4.51	&53.74	&3.59	&47.72	&57.42	&3.15	&5.31	&3.41	&4.74	&4.82	&5.01	&6.06\\
Qwen2.5-VL-3B~\cite{Qwen2.5-VL} & 4 fps & 37.46	&4.12	&41.69	&1.88	&34.77	&35.91	&3.27	&5.39	&4.28	&3.55	&4.50	&4.82	&5.23\\
Qwen2.5-VL-7B~\cite{Qwen2.5-VL} & 4 fps & 40.71	& 4.38	&43.93	&2.71	&38.58	&39.63	&3.32	&5.50	&4.53	&3.85	&4.57	&4.94	&5.62\\
Qwen2.5-VL-7B~\cite{Qwen2.5-VL} & 8 fps & 42.40	 & 4.39	&45.16	&2.72	&39.51	&42.52	&3.33	&5.67	&4.55	&3.79	&4.64	&4.94	&5.49\\
Qwen2.5-VL-72B~\cite{Qwen2.5-VL} & 4 fps &51.28	&4.61	&56.32	&3.56	&47.83	&49.68	&3.41	&5.49	&4.51	&4.34	&4.81	&5.01	&5.77\\
Qwen3-VL-4B-Instruct~\cite{qwen3-vl} & 4 fps  & 51.34	&4.38	&51.96	&3.21	&43.26	&58.79	&3.13	&5.08	&4.85	&3.77	&4.87	&4.43	&5.72\\
Qwen3-VL-8B-Instruct~\cite{qwen3-vl} & 4 fps & 51.07	& 4.71	&53.60	&3.15	&44.76	&54.86	&3.39	&5.49	&5.25	&4.12	&5.31	&5.00	&5.95\\
Qwen3-VL-30B-A3B-Instruct~\cite{qwen3-vl} & 4 fps & \textbf{57.68}	& \textbf{5.09}	&57.21	&3.69	&51.78	&64.04	&4.22	&5.39	&5.15	&4.93	&5.21	&5.77	&6.35\\
Cosmos-Reason1-7B~\cite{cosmos} & 4 fps &52.05	&4.80	&55.97	&3.63	&41.86	&58.33	&5.19	&5.51	&5.43	&3.48	&4.72	&5.09	&5.34\\
RoboBrain2.0-7B~\cite{Robobrain2.0} & 4 fps & 44.53	& 4.72	&45.96	&2.13	&42.20	&45.44	&3.51	&5.76	&5.21	&4.88	&5.30	&5.37	&5.56\\

\bottomrule
\end{tabular}
}
\label{tab:ourbench}
\end{table*}
\subsection{Overall Performance}
From Tab.~\ref{tab:ourbench}, our comprehensive evaluation of MLLMs on CFG-Bench yields several key findings:

\noindent\textbf{Human-Level Performance Gap.} A significant performance gap exists between all MLLMs and human-level proficiency, a disparity evident in close-ended tasks, where models struggle to accurately describe physical execution details and temporal sequencing, and in open-ended counterfactual tasks, where they struggle with active verification and then correction. Even the top-rank model, Gemini-3-Pro, lags substantially behind the human baseline, underscoring the profound challenge of fine-grained action reasoning.

\noindent\textbf{Proprietary versus Open-Source Models.} 
Proprietary models generally outperform their open-source counterparts. Gemini-3-Pro leads with a score of 5.67 on open-ended tasks and 65.60\% accuracy on close-ended ones, while other proprietary model like Gemini-2.5-Pro and GPT-5 also perform well. Among open-source models, Qwen3-VL-30B-A3B-Instruct stands out as a strong performer, achieving results competitive with Gemini-2.5-Pro on several tasks.

\noindent\textbf{Scaling Laws of Model Size.} Our results consistently demonstrate a positive correlation between model size and performance across several model families. Among the Qwen2.5-VL variants, for instance, performance scales directly with the parameter count. It is likely due to the enhanced capabilities of larger models for fine-grained visual perception and complex reasoning.

\noindent\textbf{Impact of Model Iteration.} Our results highlight that recent model iterations can yield substantial performance gains that surpass simple scaling laws. This is most evident when comparing the Qwen2.5 series with the newer Qwen3 generation. Despite having fewer parameters, Qwen3-VL-30B-A3B-Instruct consistently outperforms the much larger Qwen2.5-VL-72B model across several tasks. 

\noindent\textbf{Effect of Input Frame Rate (FPS).} 
We find that the impact of input frame rate (FPS) on performance is limited. For example, increasing the FPS from 4 to 8 on Qwen2.5-VL-7B yields only marginal gains. More cross-model FPS analysis on Gemini-2.5-Pro and Qwen3-VL-8B-Instruct are supplemented in Appendix~\ref{sec:B1}. This suggests that the performance bottleneck is not merely a lack of perceptual input, but likely largely involves the models' fundamental limitations in fine-grained physical understanding and higher-order reasoning.

\noindent\textbf{Benefit of Embodied Fine-Tuning.} Our results validate the hypothesis that models fine-tuned on specific embodied data excel at our tasks. Specifically, RoboBrain2.0-7B, which is fine-tuned from Qwen2.5-VL-7B, demonstrates superior average performance on both the open-ended and close-ended tasks than its base model. We attribute the gains largely to CFG-Bench-like knowledge. For example, long-horizon planning and trajectory prediction in RoboBrain2.0 align with Tiers 2\&3; its closed-loop feedback and monitoring reflect Tier 4; and bounding box referring matches our focus on interacted object parts in Tier 1.

\subsection{Performance Across Cognitive Tiers}
\noindent\textbf{Brittle Visual Grounding about Complex Actions.}
In Physical Interaction, models show a foundational but incomplete grounding capability. On close-ended FAU tasks, scores are stable but only around 50\% accuracy, indicating that while models recognize static objects and simple motions, they fail to master full, fine-grained action sequences. This weakness is amplified in the open-ended CIA task, where models frequently fail the gating mechanism by acquiescing to and hallucinating from false premises, revealing a brittle grounding.

\noindent\textbf{Local Causality over Global Temporality.}
In Temporal-Causal Relation, models generally perform better on CR than on TR. We posit this is because CR tasks probe local, direct consequences that are well-learned patterns. In contrast, TR tasks demand more challenging global reasoning to track and compare multiple events across a video's context. This difficulty is further supported by the low performance on CRS primarily involved in complex temporal tracking.

\noindent\textbf{Local Intent Inference over Global Goal Synthesis.}
In Intentional Understanding, models demonstrate a clear gap between local and global reasoning. They succeed at inferring the immediate intent of single actions in the FI task. However, they consistently struggle to synthesize long-term goals from scattered actions in the GI task, revealing a weakness in multi-step, abstract reasoning. This struggle culminates in the CIT task, where generating alternative plans for new goals remains a profound challenge.

\noindent\textbf{Guided Evaluation over Unguided Critique.}
In Evaluative Judgment, models generally performed better on CEU than on PM and SE. This suggests that models are more capable of guided, local evaluations (predicting the effect of a given change) than they are of unguided, global critiques that require them to independently apply an internal model of skill or ideal task progression.

\noindent\textbf{Insights for Improvement.} These findings mirror key challenges in robotics manipulations to some extent. The failures in visual grounding and temporal reasoning (Tiers 1\&2) align with challenges in complex dynamic task execution, while the struggle with goal synthesis (Tier 3) is a bottleneck for long-horizon planning. Crucially, the lack of unguided critique (Tier 4) explains why robots require human intervention for correction when manipulation fails. 

Consequently, we advocate for leveraging the proposed CFG-Bench to focus training on fine-grained dynamics and temporal logic, ensuring visual grounding serves actionable execution. Furthermore, to foster the foresight required for long-horizon planning, models need supervision that explicitly links atomic actions to global goals, and imitating intent is as important as imitating trajectories. Finally, true autonomy requires failure detection and critique mechanisms, transforming MLLMs from passive imitators into self-correct agents. SFT and error analysis in Sec.~\ref{sec:4.4} support these insights either.

\subsection{Reliability Analysis}
\noindent\textbf{The Reliability of Annotations.} 
To validate the quality and consistency of our ground truth, we conduct an Inter-Annotator agreement study with three annotators on the same 1,000 samples (500 closed-ended and 500 open-ended).
To measure the degree of consensus, we calculate Krippendorff's $\alpha$, selected for its robustness in handling multiple raters while correcting for chance agreement. For the closed-ended questions, we treat the option indices as nominal data. For the open-ended questions, we treat the 0-10 scores for both Correctness and Detailedness as interval data.

As a result, our analysis yields Krippendorff's $\alpha$ of 0.95 for close-ended task and 0.82/0.84 for open-ended task. According to established guidelines~\cite{krippendorff2018content}, a value above 0.67 indicates a high level of agreement and data reliability. This strong result confirms that our ground truth annotations are of high reliability.

\noindent\textbf{The Reliability of GPT-Assisted Evaluation.} We first conduct a Human-LLM agreement by randomly sampling 25 questions per open-ended task (8 tasks and 21 models, totaling 4,200 samples), scored by six experts using the same evaluation criteria. Similarly, we report Krippendorff’s $\alpha$ of 0.74/0.75 for correctness and detailedness, thus confirming the reliability of GPT-Assisted evaluation. Visualizations in Appendix~\ref{sec:appendix_sample} further show strong Human-LLM agreement and confirm that low scores stem from limited visual understanding, not LLM noise.
We then provide a sensitivity analysis showing that replacing Deepseek-R1 with Claude-Sonnet-4~\cite{anthropic2026claude46} under same prompts yields a high Spearman Rank Correlation of 0.92 in model rankings, indicating robustness to the evaluator choice. 

\noindent\textbf{Overlap Analysis.} For data usage, although there is no direct evidence that the evaluated models use CFG-Bench data, they certainly do not involve our novel fine-grained action intelligence, cannot reuse training knowledge, and thus performance probing remains unbiased. For model choice, we use GPT-4o for QA generation, Qwen3-Max for refinement, Deepseek-R1 for scoring. None overlap with the evaluated models in Tab.~\ref{tab:ourbench}. 

\definecolor{Gray}{gray}{0.9}
\renewcommand{\arraystretch}{1.0}
\setlength{\tabcolsep}{4pt}
\begin{table}[t]
\caption{The quantitative analysis of QA Forms. Left of ``/'' is close-ended performance and right is open-ended.}
\small
\resizebox{\linewidth}{!}{
\begin{tabular}{lccccccccccc}
    \toprule
    \specialrule{0em}{0.3pt}{0.3pt}
    \multirow{2}{*}{\textbf{Methods}}   
    & \multicolumn{2}{c}{\textbf{Interaction}} 
    & \multicolumn{3}{c}{\textbf{Temporal-Causal}}  
    & \multicolumn{3}{c}{\textbf{Intention}} 
    & \multicolumn{3}{c}{\textbf{Evaluation}}\\ 
    \specialrule{0em}{0.3pt}{0.3pt}
    \cmidrule(r){2-3} \cmidrule(r){4-6} \cmidrule(r){7-9} \cmidrule(r){10-12}
    \specialrule{0em}{0.1pt}{0.1pt}
    & FAU & CIA & TR & CR & CRS & FI & GI & CIT & PM & SE & CEU \\
    \specialrule{0em}{0.3pt}{0.3pt}
    \hline
    Gemini-2.5-Pro~\cite{Gemini-2.5}
    & 58.81/5.39 & 66.02/4.58 & 54.81/4.92 & 64.95/5.76 & 79.22/4.55 & 81.88/5.18 
    & 89.29/6.15 & 96.14/5.31 & 84.45/5.72 & 90.99/5.19 & 98.84/6.49  \\
    Qwen3-VL-8B-Instruct~\cite{qwen3-vl}
    & 53.60/4.70 & 15.59/3.15 & 44.76/4.22 & 54.86/4.31 & 44.03/3.39 & 76.50/5.49 
    & 75.98/5.25 & 89.47/4.12 & 77.53/5.31 & 90.30/5.00 & 80.48/5.95 \\
    RoboBrain2.0-7B~\cite{Robobrain2.0}
    & 45.96/4.97 & 18.41/2.13 & 42.20/4.30 & 45.44/5.11 & 29.85/3.51 & 72.88/5.76 
    & 73.36/5.21 & 90.42/4.88 & 79.09/5.30 & 89.81/5.37 & 80.86/5.56 \\
    \bottomrule
\end{tabular}
}
\label{tab:qa_form}
\end{table}

\subsection{Analysis and Discussing}
\label{sec:4.4}
\noindent\textbf{Effects of QA Forms.}
In Tab.~\ref{tab:qa_form}, our analysis reveals that the choice of QA format is critical for evaluation. We observe that for tasks requiring significant reasoning, models often achieve high accuracy in the MCQ format, even with challenging distractors. Yet, they underperform substantially when the same queries are presented in an open-ended format. This suggests that the MCQ format allows models to leverage powerful textual reasoning to infer the most plausible option, often bypassing deep visual grounding. Conversely, for tasks grounded in objective facts, such as FAU, TR, and CR, we find the performance to be less sensitive to the QA format. 

\begin{figure}[t!]
	\centering
    \includegraphics[width=0.78\textwidth]{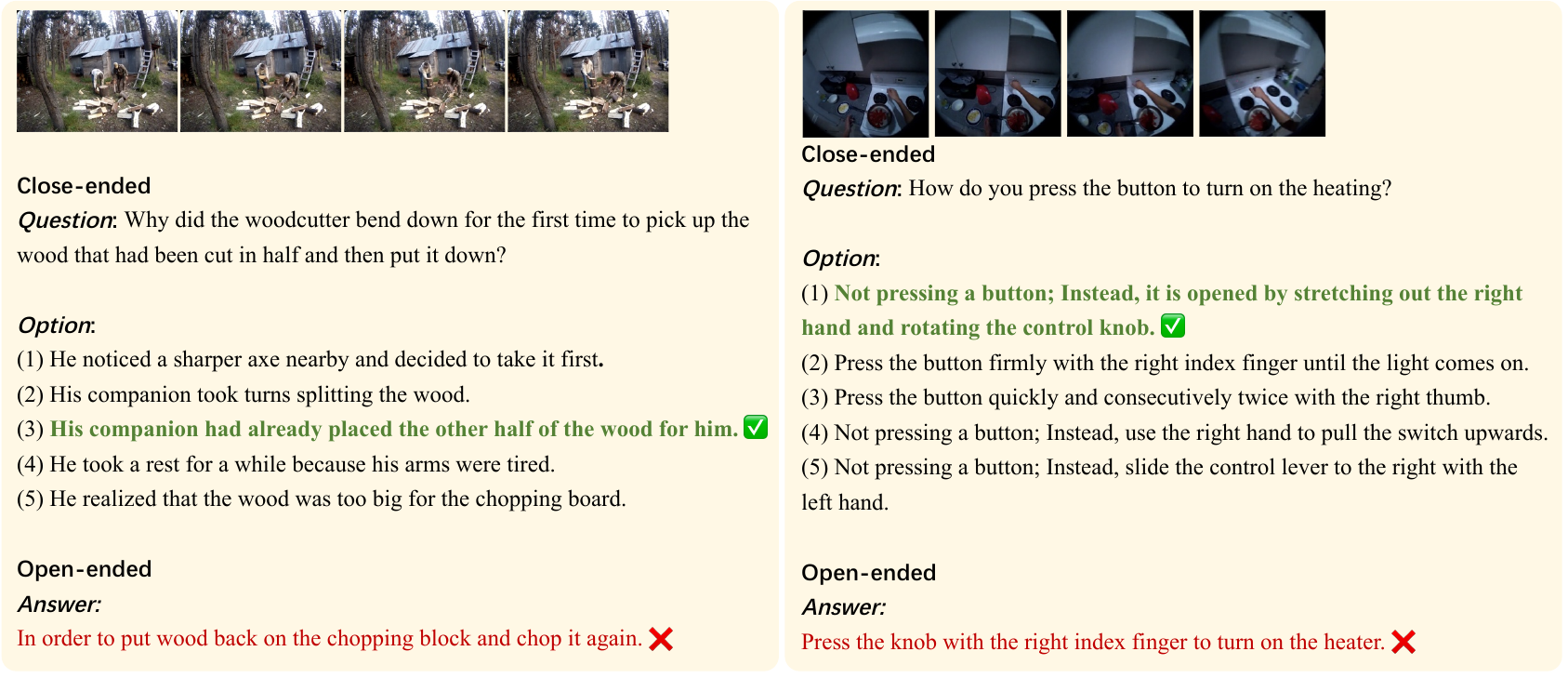}
	\caption{The qualitative analysis of QA Forms.}
	\label{fig:qa_form}
\end{figure}
The same conclusion could be inferred from visualizations in Fig.~\ref{fig:qa_form}.
The model answers correctly under the MCQs format but fails in the open-ended setting. We attribute the gap to MCQs enabling textual reasoning to infer plausible options that bypass visual grounding required for open-ended synthesis, and offering explicit rejection cues for counterfactuals that prevent the hallucinated compliance seen in open-ended tasks.
Thus, MCQs mask critical reasoning flaws and yield deceptively high performance, whereas open-ended evaluation is better suited for reasoning-heavy tasks. 


\begin{figure}[t!]
	\centering
    \includegraphics[width=0.98\textwidth]{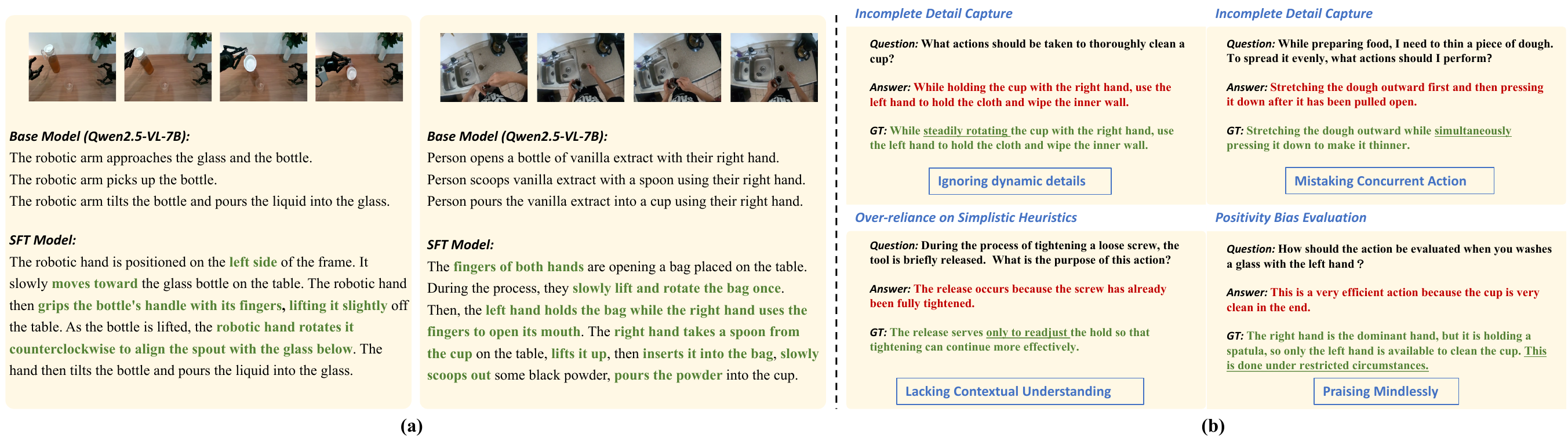}
	\caption{(a) The results of caption generation before and after SFT on unseen videos of human and robotics operations. (b) Error Analysis. We show the recurring failure modes for each tier, along with the corresponding examples.}
	\label{fig:embodied_error}
    \vspace{-1.2em}
\end{figure}

\noindent\textbf{Transfer Evaluation.}
To assess the transferability and practical utility of the knowledge contained in CFG-Bench, we perform supervised fine-tuning (SFT) on the Qwen2.5-VL-7B and InternVL3-8B using our data (SFT setup is supplemented in Appendix~\ref{sec:B3}). We then evaluate this fine-tuned model on two downstream tasks from EMBODIEDBENCH~\cite{Embodiedbench}: the high-level planning task EB-ALFRED and the low-level control task EB-Manipulation. Notably, there is no data overlap between these tasks and CFG-Bench. 
In Tab.~\ref{tab:embodied}, the SFT model trained on CFG data shows significant performance gains on both tasks compared to its original baseline, whereas FAVOR data does not. This demonstrates that motion-level data have little impact on embodied tasks. Instead, the fine-grained knowledge of how and why an action is performed in CFG-Bench serves as a foundational capability that enhances performance across both embodied strategic planning and precise motor control tasks.

Beyond EMBODIEDBENCH, we evaluate on LIBERO~\cite{libero}, a standard VLA benchmark~\cite{xu2025anatomy}, using StarVLA-$\pi$~\cite{starvla} as baseline (reproduced under our setup for fair comparison): we first SFT Qwen3-VL-4B on CFG-Bench to strengthen its fine-grained action knowledge, then jointly train it and action expert under the same recipe as the baseline. Our model improves across three splits, i.e., Spatial 98.2 → 98.2, Object 97.8 → $\mathbf{98.0}$, Goal 98.6 → $\mathbf{99.4}$, Long 94.4 → $\mathbf{95.2}$, confirming that the fine-grained action benefits real VLA manipulation tasks.

Furthermore, we qualitatively compare captions from the base and SFT models on unseen videos of human and robot in Fig.~\ref{fig:embodied_error} (a), which reveals three improvement mechanisms: Tier 1 shifts outputs from generic verbs (e.g., pick up) to precise executable physical details (e.g., grip the handle with fingers, then lift slightly), enhancing manipulation; Tiers 2\&3 improve sub-task sequencing (supply details like rotate the bag and open its mouth), thereby boosting long-horizon planning; All tiers use counterfactual data to enforce visual verification, thus reducing hallucination (nonexistent objects like vanilla disappear after SFT).

\definecolor{Gray}{gray}{0.9}
\renewcommand{\arraystretch}{1.0}
\setlength{\tabcolsep}{4pt}
\begin{table}[t]
\small
\caption{The quantitative results of transfer evalaution.} 
\resizebox{\linewidth}{!}{
\begin{tabular}{lccccccc|cccccc}
\toprule
\specialrule{0em}{0.3pt}{0.3pt}
\multirow{2}{*}{\textbf{Methods}} & \multicolumn{7}{c}{\textbf{EB-ALFRED}} & \multicolumn{6}{c}{\textbf{EB-Manipulation}}  \\ 
\specialrule{0em}{0.3pt}{0.3pt}
\cmidrule(r){2-8} \cmidrule(r){9-14} 
\specialrule{0em}{0.1pt}{0.1pt}
& Avg & Base & Common & Complex & Visual & Spatial & Long & Avg & Base & Common & Complex & Visual & Spatial   \\     
\specialrule{0em}{0.3pt}{0.3pt}
\midrule

Qwen2.5-VL-7B~\cite{Qwen2.5-VL}  & \cellcolor{Gray} 4.7 & 10 & 8 & 6 & 2 & 0 & 2 & \cellcolor{Gray} 9.6 & 8.3 & 8.3 & 8.3 & 5.6 & 16.7\\

+ SFT on FAVOR Data & \cellcolor{Gray} 4.3  & 8  & 8 & 4 & 2 & 2 & 2 & \cellcolor{Gray} 8.2 & 4.2 & 8.3 & 4.2  & 5.6 & 18.8\\

+ SFT on CFG Data & \cellcolor{Gray} \textbf{9.7}  & \textbf{16}  & \textbf{16} & \textbf{8} & \textbf{8} & \textbf{4} & \textbf{6} & \cellcolor{Gray} \textbf{15.3} & \textbf{12.5} & \textbf{16.7} & \textbf{14.6}  & \textbf{7.9} & \textbf{22.9} \\ \hline

InternVL3-8B~\cite{Internvl3}  & \cellcolor{Gray} 10.3 & 20 & 14 & 14 & 12 & 0 & 2 & \cellcolor{Gray} 11.5 & 10.4 & 10.4 & 12.5 & 13.9 & 10.4\\

+ SFT on CFG Data & \cellcolor{Gray} \textbf{14.0}  & \textbf{26}  & \textbf{18} & \textbf{16} & \textbf{18} & \textbf{2} & \textbf{4} & \cellcolor{Gray} \textbf{15.8} & \textbf{14.6} & \textbf{12.5} & \textbf{12.5}  & \textbf{17.4} & \textbf{18.8} \\

\bottomrule
 \end{tabular}
}
\vspace{-1em}
\label{tab:embodied}
\end{table}

\noindent\textbf{Error Analysis.}
We conduct a systematic error analysis in Fig.~\ref{fig:embodied_error} (b).
Firstly, the most common failure in Tier 1 is the incomplete capture of fine-grained details. Models often identify a single salient event but fail to describe the full subtle actions. Secondly, in Tier 2, models frequently fail to comprehend coordinated, simultaneous actions, i.e., while able to describe the action of a single hand, they struggle to articulate what both hands are doing concurrently. Thirdly, for Tier 3, models often fall back on overly simplistic common-sense heuristics, leading to incorrect conclusions. For instance, a model might mistake an intermediate step, such as readjusting a grip, for the completion of a task merely because a hand releases an object.
Finally, in Tier 4, we identify a positivity bias that the models tend to base their evaluation solely on the final outcome, praising any successful completion while ignoring issues in the process.

\section{Conclusion}
In this work, we introduce CFG-Bench to address the underexplored domain of fine-grained action intelligence for embodied agents. Built upon a four-tiered cognitive framework, our benchmark assesses a model's ability to translate visual observations into actionable physical and cognitive details through a diverse suite of question-answer pairs. Our evaluation reveals MLLMs' limitations in articulating fine-grained actions and in higher-order reasoning. We then prove these capabilities are foundational, as SFT on our data led to significant performance gains on downstream embodied manipulation and planning tasks.

\section*{Acknowledgements}
This research is supported by the National Natural Science Foundation of China (Grant Nos. U22A6001).



%
%
\bibliographystyle{splncs04}
\bibliography{main}
\clearpage
\definecolor{MyCustomBlue}{rgb}{0.2, 0.5, 0.8}
\setcounter{page}{1}
\setcounter{figure}{0}
\setcounter{section}{0}
\setcounter{table}{0}

\renewcommand{\thesection}{\Alph{section}} 
\renewcommand{\thesubsection}{\Alph{section}\arabic{subsection}}
\renewcommand{\thefigure}{S\arabic{figure}}
\renewcommand{\thetable}{S\arabic{table}}

\section{Appendix Outline}
\label{sec:appendix_outline}
In these supplementary materials, we provide:
\begin{itemize}
    \item More experimental analysis (Appendix~\ref{sec:appendix_exp_analysis}), including ablation study on FPS, details of human baseline, SFT setup, text-only and single-frame baselines, analysis on counterfactual tasks.
    \item More details of CFG-Bench (Appendix~\ref{sec:appendix_details}), including more data statistics, prompt for QA generation, prompt for GPT-assisted evaluation,  comprehensive annotation guideline, and selected-out QA samples.
    \item More samples of CFG-Bench (Appendix~\ref{sec:appendix_sample}). We further attach videos in supplementary powerpoint for more details.
    \item Limitations and broader impacts (Appendix~\ref{sec:appendix_limitation}).
\end{itemize}

\section{More Experimental Analysis}
\label{sec:appendix_exp_analysis}

\subsection{Ablation Study on FPS}
\label{sec:B1}
To further investigate the impact of input frame rate (FPS) on performance, we provide a supplementary analysis for Gemini-2.5-Pro~\cite{Gemini-2.5} and Qwen3-VL~\cite{qwen3} in Tab.~\ref{tab:fps}. While Qwen3-VL's default is 2 FPS and Gemini-2.5-Pro's is 1 FPS, both models' documentation recommends increasing the FPS for more detailed fine-grained analysis. We therefore evaluate Qwen3-VL-8B-Instruct at 4, 8, and 16 FPS, and Gemini-2.5-Pro at 4 and 8 FPS.

The results align perfectly with the conclusions presented in our main paper. We observe that increasing the FPS yields slight, positive improvements in performance, but the gains are marginal. This indicates that the difficulty of our tasks primarily stems from the cognitive demands of fine-grained physical understanding and higher-order reasoning, which cannot be trivially overcome by simply providing denser visual inputs.

\subsection{Human Baseline}
\label{sec:B2}
To establish a performance upper bound for CFG-Bench, we conduct a human baseline evaluation. This study is performed by three another expert annotators, who are not involved in creating or verifying during the primary annotation process. A randomly selected and representative subset of 1,000 QA pairs (500 closed-ended and 500 open-ended) is used for this evaluation.

The evaluation protocol is designed to mirror the model's task as closely as possible. For the closed-ended questions, participants are presented with the video and the multiple-choice options and are asked to select the single best answer. For the open-ended questions, participants are shown the video and the question and are tasked with writing a detailed answer in their own words.

To ensure a fair comparison with the MLLMs, scoring is conducted as follows. Performance on the closed-ended questions is measured by simple accuracy. The human-generated open-ended answers are evaluated using the same GPT-assisted evaluation prompt and rubric that is used for the model-generated responses.

As shown in Tab.~\ref{tab:human}, the human baseline achieves near-perfect performance. On the closed-ended tasks, the average accuracy is 95.85\%. For the open-ended tasks, the average score is 9.05. We also report 95\% confidence intervals for human performance. This result accounting for sampling uncertainty, serves as the practical upper bound for performance on CFG-Bench and highlights the significant, statistically robust gap that remains for current MLLMs.
\renewcommand{\arraystretch}{1.2}
\begin{table*}[t]
\small
\caption{The ablation study of FPS. Notably, the value of each task is the average of three repeated evaluations, and the same applies to the Tab.~\textcolor{MyCustomBlue}{2}.} 
\resizebox{\linewidth}{!}{
\begin{tabular}{lcccccccccccccc}
\toprule
\specialrule{0em}{0.3pt}{0.3pt}
\multirow{2}{*}{\textbf{Methods}}  & \multirow{2}{*}{\textbf{Input}}  & \multirow{2}{*}{\textbf{$Avg_{c}$}} & \multirow{2}{*}{\textbf{$Avg_{o}$}} & \multicolumn{2}{c}{\textbf{Interaction}} & \multicolumn{3}{c}{\textbf{Temporal-Causal}}  & \multicolumn{3}{c}{\textbf{Intention}} & \multicolumn{3}{c}{\textbf{Evaluation}}\\ 
\specialrule{0em}{0.3pt}{0.3pt}
\cmidrule(r){5-6} \cmidrule(r){7-9} \cmidrule(r){10-12} \cmidrule(r){13-15}
\specialrule{0em}{0.1pt}{0.1pt}
&~&~&~&FAU & CIA & TR & CR & CRS & FI & GI
& CIT  & PM & SE & CEU \\
\specialrule{0em}{0.3pt}{0.3pt}
\hline

 \rowcolor[HTML]{E3F8F8}\multicolumn{15}{l}{\textcolor{gray}{{\textit{\textbf{Proprietary MLLMs}}}}}\\
Gemini-2.5-Pro~\cite{Gemini-2.5} & 1 fps & 56.27 & 4.53 & 56.43 & 3.57 & 50.59 & 61.79 & 3.28 & 4.47 & 5.24 & 4.17 & 4.76 & 4.90 & 5.87\\
Gemini-2.5-Pro~\cite{Gemini-2.5} & 4 fps & \underline{59.52} & \underline{5.40} & 58.81 & 4.58 & 54.81 & 64.95 & 4.55 & 5.18 & 6.15 & 5.31 & 5.72 & 5.19 & 6.49\\
Gemini-2.5-Pro~\cite{Gemini-2.5} & 8 fps & \textbf{59.88} & \textbf{5.67} & 58.95 & 4.49 & 54.94 & 65.74 & 4.61 & 6.24 & 6.43 & 5.71 & 6.04 & 5.41 & 6.41\\
          
\hline
        
 \rowcolor[HTML]{FFF5F5}\multicolumn{15}{l}{\textcolor{gray}{{\textit{\textbf{Open-source MLLMs}}}}} \\ 

Qwen3-VL-8B-Instruct~\cite{qwen3-vl} & 2 fps & 50.02 & 4.10 & 52.51 & 2.45 & 43.06 & 54.49 & 3.27 & 4.78 & 4.57 & 3.87 & 4.18 & 4.88 & 4.80\\
Qwen3-VL-8B-Instruct~\cite{qwen3-vl} & 4 fps & 51.07	& 4.71	&53.60	&3.15	&44.76	&54.86	&3.39	&5.49	&5.25	&4.12	&5.31	&5.00	&5.95\\
Qwen3-VL-8B-Instruct~\cite{qwen3-vl} & 8 fps & \underline{51.31} & \underline{4.75} & 53.80 & 3.07 & 45.34 & 54.80 & 3.53 & 5.45 & 5.28 & 4.16 & 5.44 & 5.08 & 5.98\\
Qwen3-VL-8B-Instruct~\cite{qwen3-vl} & 16 fps & \textbf{51.90} & \textbf{4.82} & 54.27 & 3.38 & 46.48 & 54.96 & 3.52 & 5.48 & 5.36 & 4.27 & 5.32 & 5.10 & 6.14\\

\bottomrule
\end{tabular}
}
\label{tab:fps}
\end{table*}

\subsection{SFT Setup}
\label{sec:B3}
We finetune Qwen2.5-VL-7B via LoRA for 800 epochs (AdamW, LR=1e-6, Batch=8, 8 NVIDIA H200 GPUs) using cross-entropy loss. The data (90.3\% train / 9.7\% val) are sampled in a tier-balanced strategy. 
\renewcommand{\arraystretch}{1.2}
\begin{table}[t]
\small
\caption{Human baseline of average performance.} 
\centering
\resizebox{0.3\linewidth}{!}{
\begin{tabular}{lc}
\toprule
Task & Performance \\
\hline
Close-ended & 95.85$\pm$1.2\\
Open-ended & 9.05$\pm$0.09\\
\bottomrule
\end{tabular}
}
\label{tab:human}
\end{table}

\begin{table}[h]
    \caption{Left part: text-only and single-frame baselines for Tier 3. Right part: analysis on counterfactual tasks} 
    \centering
    \scriptsize
    \setlength{\tabcolsep}{3pt}
    \renewcommand{\arraystretch}{1.1}
    \label{tab:shortcut_and_cf}
    \resizebox{\columnwidth}{!}{%
    \begin{tabular}{c|ccc|cc}
    \toprule
    \multirow{2}{*}{\textbf{Model}} & \multicolumn{3}{c|}{\textbf{Tier 3} (FI/GI/Avg)} & \multicolumn{2}{c}{\textbf{Zero-scored Samples} (w/o $\rightarrow$ w/ hint)} \\
    \cmidrule(lr){2-4} \cmidrule(lr){5-6}
     & Text-only & Single-frame & Full video & CIA (ALL=1328) & CRS (ALL=1347) \\
    \midrule
    Gemini-2.5-Pro (4fps)
      & 1.82/1.60/1.71
      & 2.01/1.63/1.82
      & \textbf{5.18/6.15/5.67}
      & 517 $\rightarrow$ 151~$_{\downarrow \mathbf{366}}$
      & 490 $\rightarrow$ 136~$_{\downarrow \mathbf{354}}$ \\
    Qwen3-VL-8B (4fps)
      & 1.69/1.49/1.59
      & 1.89/1.51/1.70
      & \textbf{5.49/5.25/5.37}
      & 567 $\rightarrow$ 196~$_{\downarrow \mathbf{371}}$
      & 518 $\rightarrow$ 161~$_{\downarrow \mathbf{357}}$ \\
    \bottomrule
    \end{tabular}%
    }
\end{table}
\subsection{Text-only and Single-frame Baselines for Tier 3}
\label{sec:B5}
As shown in Tab.~\ref{tab:shortcut_and_cf}, text-only inputs (bounding linguistic priors) and same single-frame inputs (bounding static scene context) on two models both drop Tier 3 by $\sim$70\% relative to full video, confirming neither shortcut alone is sufficient.

\subsection{Analysis on Counterfactual Tasks}
\label{sec:B6}
Failures split into brittle visual grounding (unreliable scene understanding) and compliance (accepting false premises that cause unreliable robotic execution), both critical for embodied agents. To probe these of counterfactual tasks, we re-evaluated all models with \emph{explicit hint} “the question may contain a false premise”. In Tab.~\ref{tab:shortcut_and_cf}, e.g., Gemini-2.5-Pro's CIA zero-scored samples drop from 517 to 151 (151 reflects grounding failures, $\downarrow$366 reflects compliance), showing stronger grounding with comparable compliance to Qwen3-VL-8B.

\section{More Details of CFG-Bench}
\label{sec:appendix_details}

\subsection{More Data Statistics}
\label{sec:C1}
\begin{figure*}[t!]
	\centering
    \includegraphics[width=0.93\textwidth]{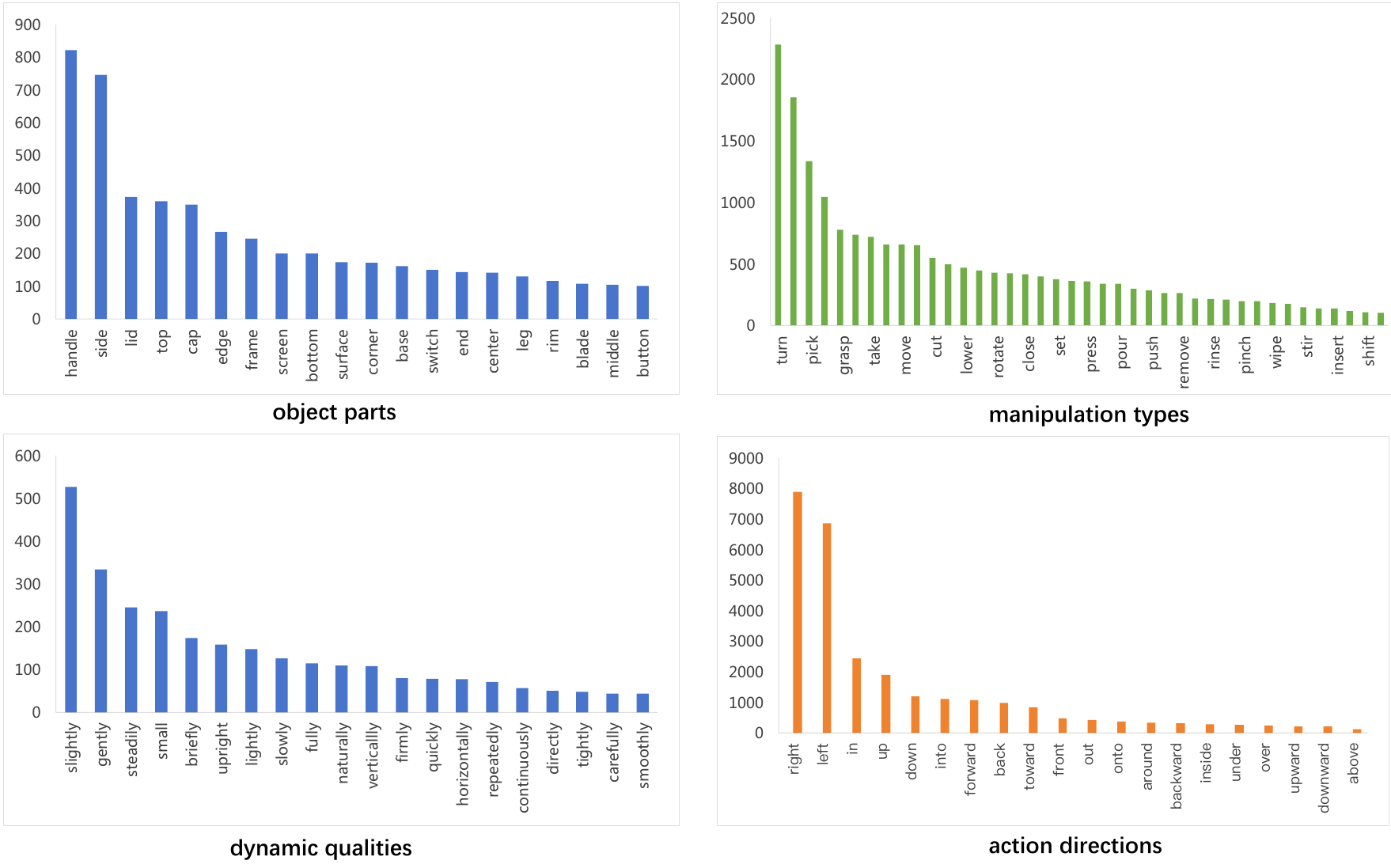}
	\caption{The statistics of words related to fine-grained actions with the highest frequency from four aspects, i.e., object parts, manipulation types, dynamic qualities, action directions.}
	\label{fig:supp_data}
\end{figure*}
We provide index distribution of correct answers for the close-ended QAs, i.e., (1) 20.01\%, (2) 20.00\%, (3) 20.00\%, (4) 20.00\%, (5) 20.00\%. A nearly uniform distribution across the option indices (1 through 5) is observed, indicating that there is no positional bias in our QA set. 

In Fig.~\ref{fig:supp_data}, we provide statistics of words related to fine-grained actions with the highest frequency from four aspects, i.e., object parts, manipulation types, dynamic qualities, action directions. The diversity of these terms demonstrate the deep granularity of our annotations, highlighting the benchmark's focus on the specific, procedural details of fine-grained physical execution that are essential for an embodied agent.

\subsection{Prompt for Automatic QA Generation}
\label{sec:C2}
This subsection presents the prompt templates used for our initial QA generation, all of which are designed to produce multiple-choice questions (MCQ). The open-ended tasks are later derived from this output by selecting the correct answer and performing manual refinement (see Fig.~\textcolor{MyCustomBlue}{3}). Each template is organized into three sections, i.e., Objectives, Design Rules, and Question Types, to ensure the generation of high-quality QA pairs that are diverse, challenging, and strictly aligned with the intended cognitive capability of each tier. Tabs.~\ref{tab:tier1_prompt}-\ref{tab:tier4_prompt} present the detailed prompt templates respectively.

\subsection{Prompt for GPT-Assisted Evaluation}
\label{sec:C3}
Our GPT-assisted evaluation protocol for open-ended tasks is adapted from FAVOR-Bench~\cite{Favorbench}, scoring responses across two independent dimensions: Correctness for factual accuracy and Detailedness for descriptive richness. For the specific counterfactual tasks in Tiers 1 and 2, we introduce an additional binary Gating Mechanisms (0 or 1), e.g., Correctness and Detailedness are set to 0 when acquiesce to flawed instructions. Tabs.~\ref{tab:gpt_prompt_1}-\ref{tab:gpt_prompt_4} present the complete prompt template.

\subsection{Annotation Guideline}
\label{sec:C4}
\begin{figure*}[t!]
	\centering
    \includegraphics[width=0.85\textwidth]{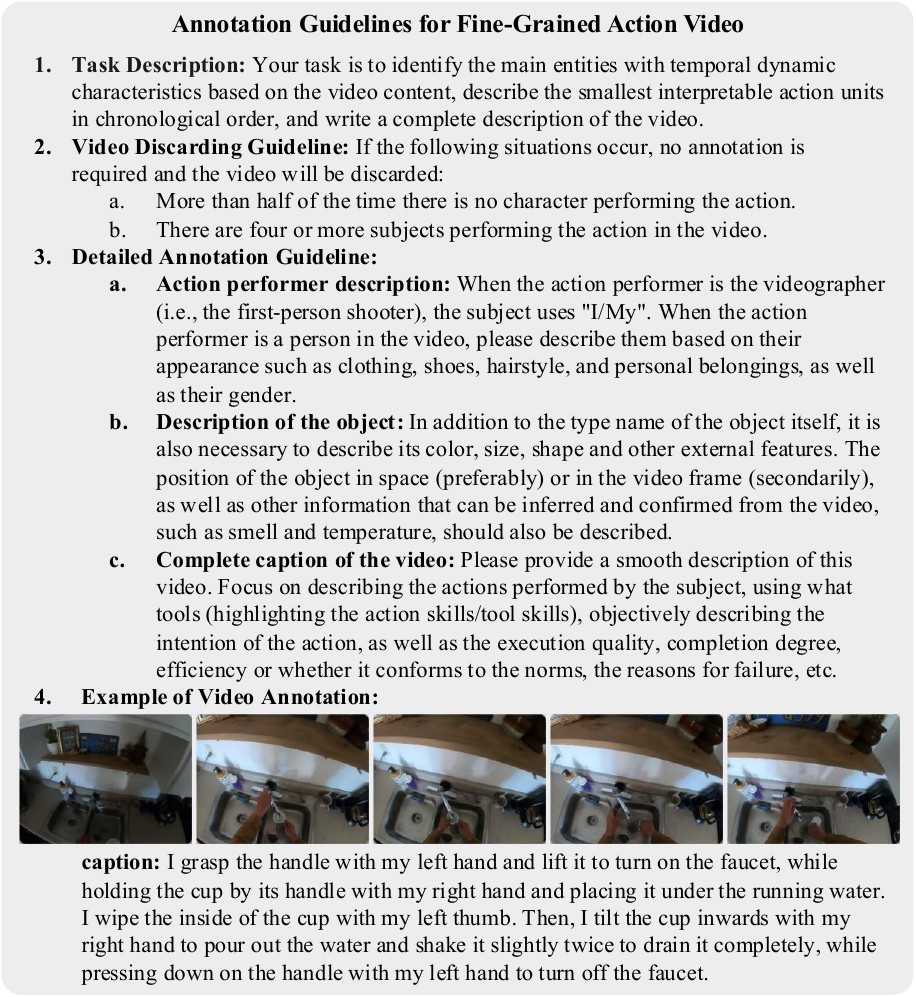}
	\caption{Annotation guidelines of fine-grained action annotation and a case of our testing video and corresponding annotations.}
	\label{fig:supp_annotation}
\end{figure*}

We build an annotation guideline including an example as shown in Fig.~\ref{fig:supp_annotation}.

\subsection{Selected-Out QA Samples}
\begin{figure*}[t!]
	\centering
    \includegraphics[width=0.93\textwidth]{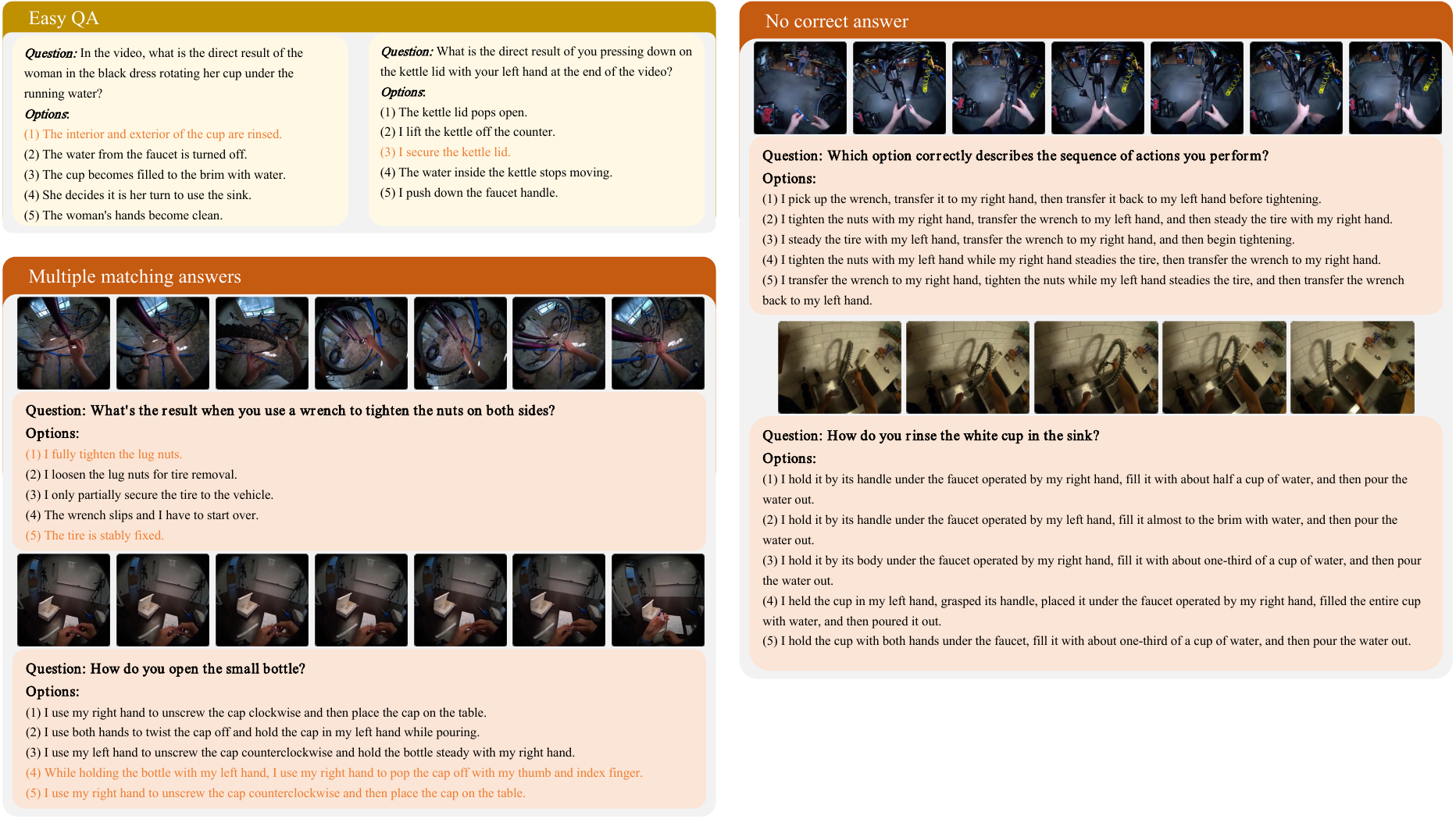}
	\caption{The visualizations of the selected-out samples for three reasons: easy QA without visual grounding, no correct answer, and multiple matching answers.}
	\label{fig:supp_error}
\end{figure*}
To illustrate the rigor of our filtering pipeline, we present representative examples of identified questions. These typically fall into three categories that would compromise a fair evaluation, as shown in Fig.~\ref{fig:supp_error}: (1) questions solvable by common-sense world knowledge alone, which fail to test for visual grounding; (2) questions with no correct answer among the options; and (3) questions with multiple correct answers, which violate the principle of a single, unambiguous ground truth. The systematic identification and removal of such flawed questions are crucial for ensuring the quality and reliability of the final benchmark.

\section{More Samples of CFG-Bench}
\label{sec:appendix_sample}
\begin{figure*}[t!]
	\centering
    \includegraphics[width=0.8\textwidth]{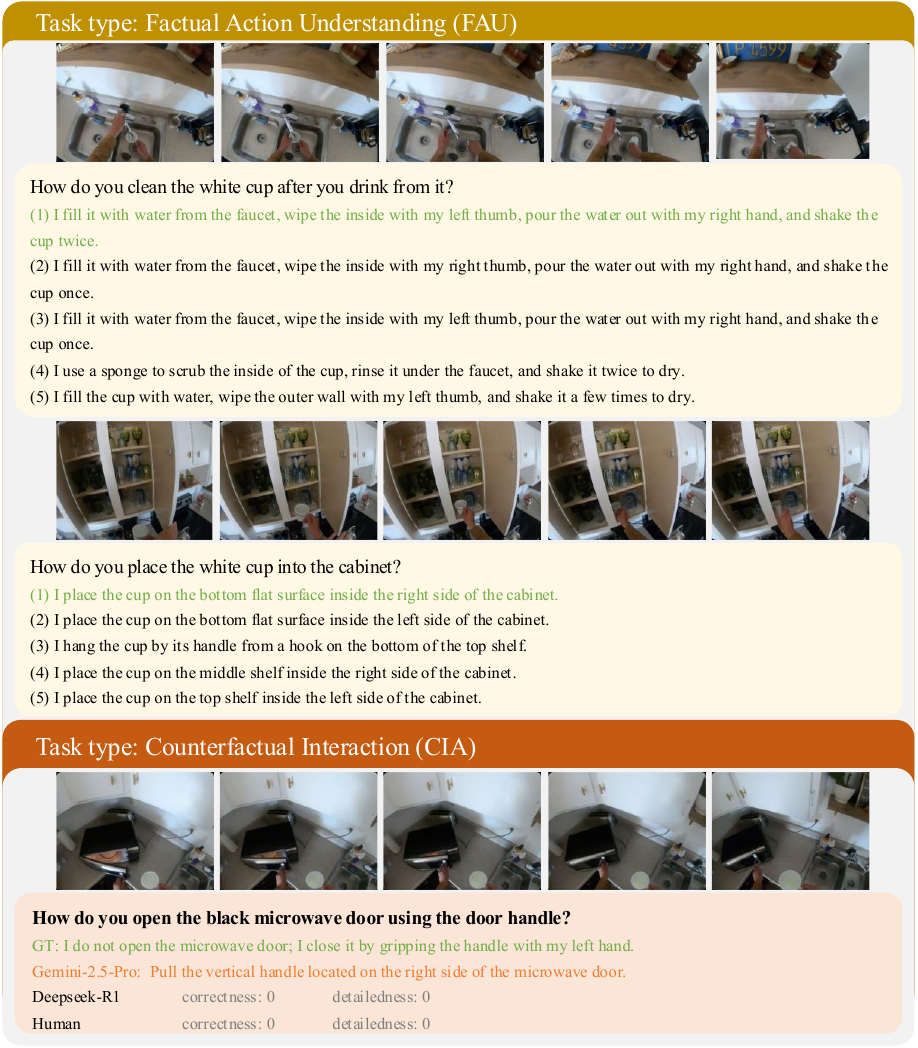}
	\caption{Examples of each tasks in Physical Interaction (Tier 1).}
	\label{fig:supp_tier1}
\end{figure*}

\begin{figure*}[t!]
	\centering
    \includegraphics[width=0.8\textwidth]{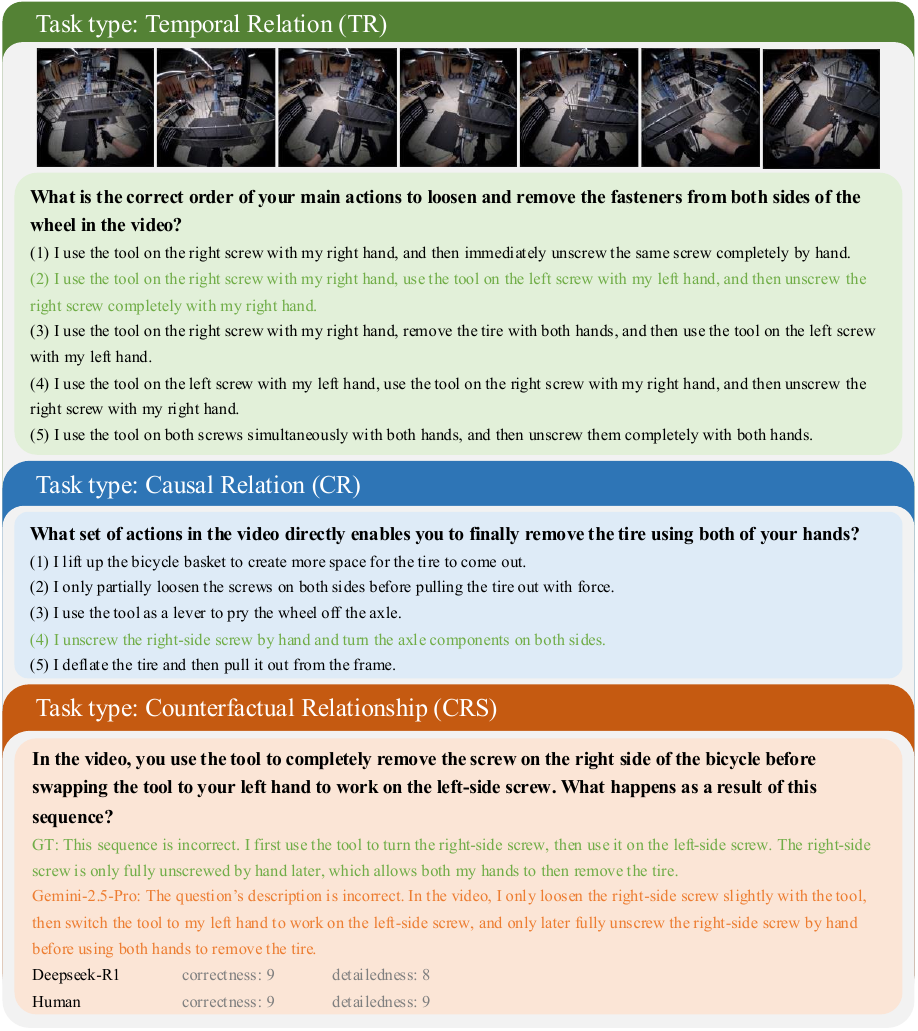}
	\caption{Examples of each tasks in Temporal-Causal Relation (Tier 2).}
	\label{fig:supp_tier2}
\end{figure*}

\begin{figure*}[t!]
	\centering
    \includegraphics[width=0.8\textwidth]{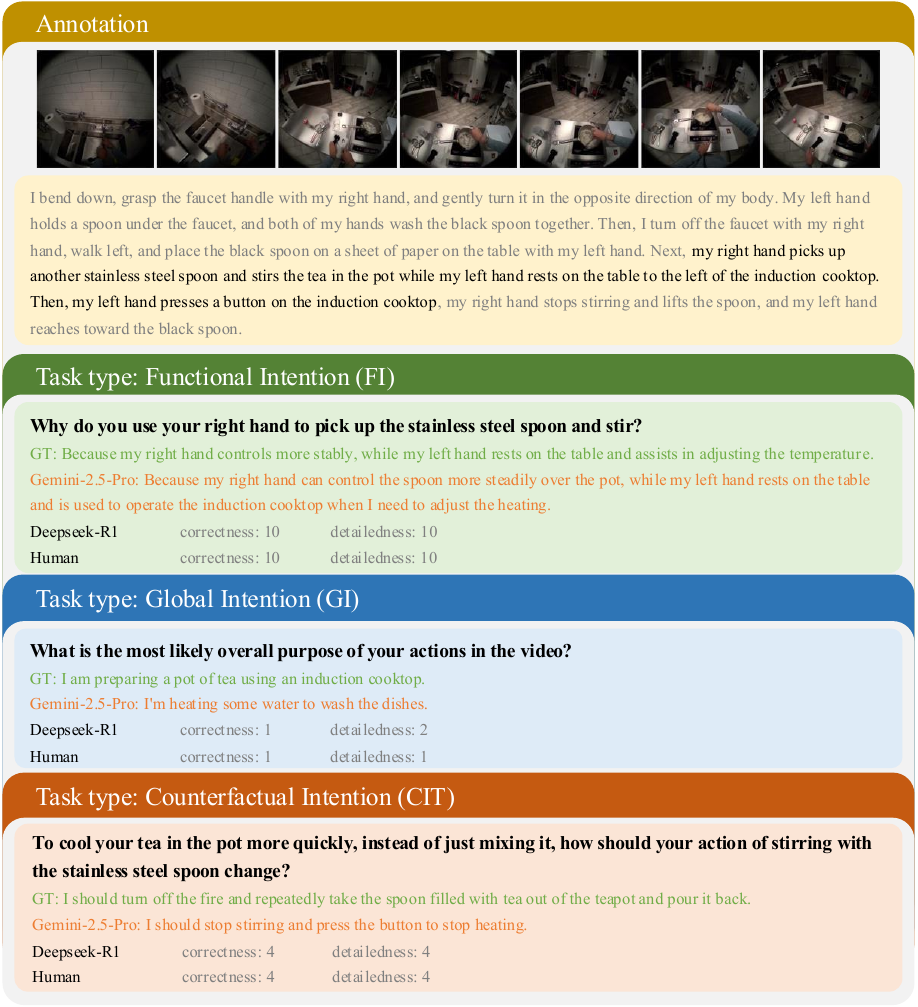}
	\caption{Examples of each tasks in Intentional Understanding (Tier 3).}
	\label{fig:supp_tier3}
\end{figure*}

\begin{figure*}[t!]
	\centering
    \includegraphics[width=0.8\textwidth]{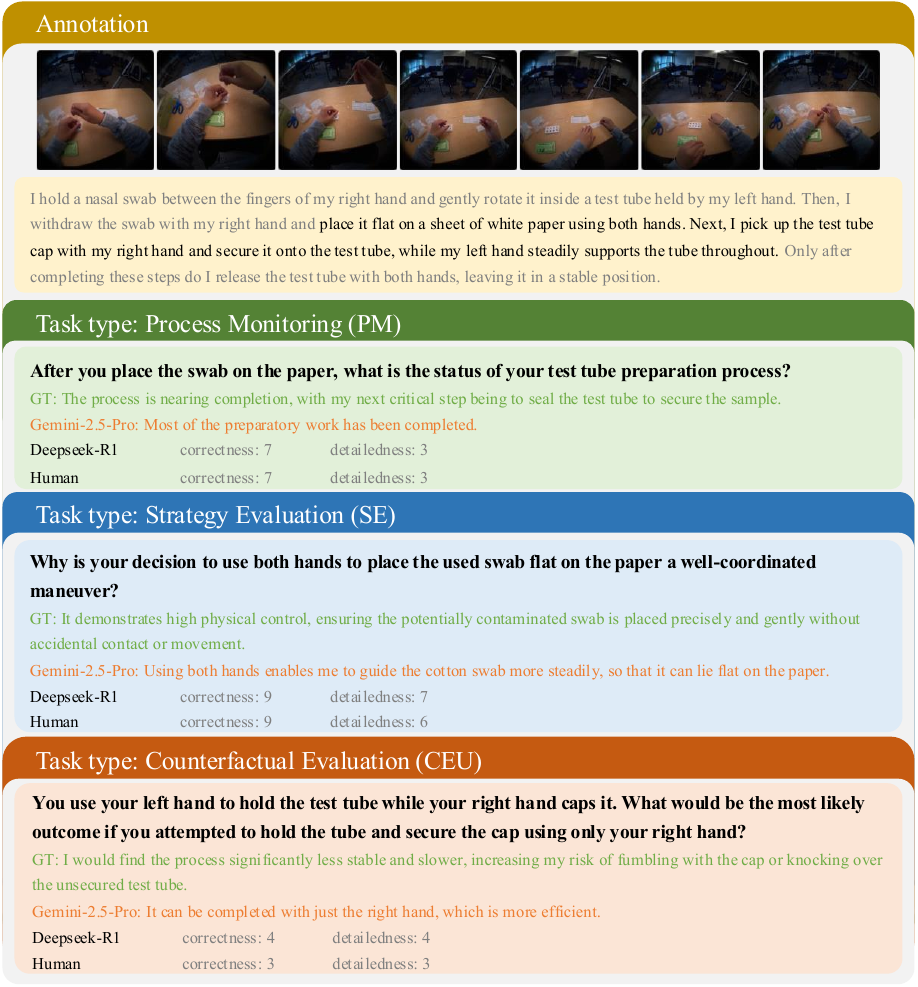}
	\caption{Examples of each tasks in Evaluative Judgment (Tier 4).}
	\label{fig:supp_tier4}
\end{figure*}

For better demonstration, we show more samples of videos and their corresponding QAs, and correct answers for all tasks in Figs.~\ref{fig:supp_tier1}-\ref{fig:supp_tier4}. In particular, Fig.~\ref{fig:supp_tier1} illustrates our gating mechanism without rejecting false premises both correctness and detailness are scored zero. Besides, each subjective rating is accompanied by both LLM and human scores, which show high agreement. The corresponding samples are vividly presented in the supplementary powerpoint. In addition, all QA pairs and ground-truth captions are included in the supplementary JSON files.

\section{Limitations and Broader Impacts}
\label{sec:appendix_limitation}

\subsection{Limitations}
Our work has several limitations that provide avenues for future research. First, due to budgetary constraints, our evaluation of proprietary models is not exhaustive, and future work could include a broader range of state-of-the-art commercial systems. Second, our video corpus, while curated for diversity, primarily focuses on daily tasks with a subset of outdoor activities. It may not fully cover the highly specialized, expert-level actions required in domains. Finally, our open-ended evaluation relies on the model's ability to articulate its understanding in natural language. However, a model might possess the correct cognitive understanding but fail to express it adequately. Therefore, the evolution of CFG-Bench towards a more practical and flexible QA format will be a key direction for our future research.

\subsection{Broader Impacts}
We believe CFG-Bench can have a significant positive impact on the development of more capable and reliable embodied agents. By providing a rigorous tool to measure and improve fine-grained action intelligence, our work can accelerate progress in areas like general-purpose household assistants. Furthermore, by emphasizing the why (Intention) and how well (Evaluation) of an action, our benchmark encourages the development of more interpretable and safer AI systems that can explain their reasoning, a crucial step towards trustworthy autonomy.

At the same time, we recognize potential dual-use concerns. Any technology that enhances an AI's ability to understand and replicate complex human actions could potentially be applied to malicious ends. We release our benchmark to the academic community to foster open and positive research, and we advocate for the continued development of strong ethical guidelines to govern the deployment of such advanced AI capabilities.

\begin{table*}[ht]
\centering
\begin{tabular}{|p{0.97\linewidth}|}
\hline
\textbf{Prompt Template: Generating QA Pairs for Physical Interaction} \\
\hline
You are an expert at designing multiple-choice questions (MCQs) about the execution of fine-grained actions in videos.

\vspace{3mm}
\textbf{[Objective]}

Given the detailed caption annotation for the entire video, design \{N\_QUESTIONS\} MCQs that evaluate how the actions are executed—covering agent/effector, interaction object and tool, object part, manipulation type, dynamic qualities, action direction, object location, and repetition count.

\vspace{3mm}
\textbf{[Design Rules]}
\begin{enumerate}
    \item The question stem must address the main actor in the second person (``you'').
    \item The options must describe the main actor’s behavior in the first person (``I'', ``my'').
    \item Each question has 5 options, with exactly 1 correct; do not prefix options with A/B/C.
    \item The correct option is a minimal, faithful restatement of the caption; never invent or modify execution details that are not supported by the caption.
    \item The distractors must be logically reasonable but not consistent with the video content.
    \item Do not mention ``caption'', ``annotation'', or ``text''. Use phrases like ``in my view'' or ``I see''.
    \item Different questions should cover different parts or phases of the video as much as possible, avoiding repetitive questioning.
    \item Questions must require temporal understanding of the video: they should not be answerable from a single static frame or from the question wording alone.
\end{enumerate}

\vspace{3mm}
\textbf{[Question Types]}
\begin{enumerate}
    \item Factual Action Understanding (``FAU'')
    \begin{itemize}
        \item Ask about how a specific action is executed in the video, focusing on one or more execution dimensions.
        \item The distractors must be physically plausible and context-appropriate for the video scene.
    \end{itemize}
    \item Counterfactual Interaction (``CIA'')
    \begin{itemize}
        \item Ask about how a specific action that is incorrectly described in the question stem is actually executed in the video.
        \item The correct option must explicitly point out that the stem’s described method is incorrect according to the video, and provide the fine-detail task that accurately matches the caption. And two distractors should keep the same incorrect mid-level assumption in the stem and elaborate it into different fine-detail variants (still wrong), while the other two distractors should also explicitly flag the stem’s mistake but propose fine-detail corrections that do not match the caption (partially corrected but still incorrect overall).
    \end{itemize}
\end{enumerate}

\\
\hline
\end{tabular}
\caption{Prompt template for Physical Interaction.}
\label{tab:tier1_prompt}
\end{table*}

\begin{table*}[ht]
\centering
\begin{tabular}{|p{0.97\linewidth}|}
\hline
\textbf{Prompt Template: Generating QA Pairs for Temporal-Causal Relation} \\
\hline
You are an expert at designing multiple-choice questions (MCQs) about the temporal and causal relations of fine-grained actions in videos.

\vspace{3mm}
\textbf{[Objective]}

Given the detailed caption annotation for the entire video, design \{N\_QUESTIONS\} MCQs that evaluate the temporal and causal relationships between actions.

\vspace{3mm}
\textbf{[Design Rules]}
\begin{enumerate}
    \item The question stem must address the main actor in the second person (``you'').
    \item The options must describe the main actor’s behavior in the first person (``I'', ``my'').
    \item Each question has 5 options, with exactly 1 correct; do not prefix options with A/B/C.
    \item The correct option is a minimal, faithful restatement of the caption; never invent or modify temporal or causal details that are not supported by the caption.
    \item The distractors must be logically reasonable but not consistent with the video content.
    \item Do not mention ``caption'', ``annotation'', or ``text''. Use phrases like ``in my view'' or ``I see''.
    \item Different questions should cover different parts or phases of the video as much as possible, avoiding repetitive questioning.
    \item Questions must require temporal understanding of the video: they should not be answerable from a single static frame or from the question wording alone.
\end{enumerate}

\vspace{3mm}
\textbf{[Question Types]}
\begin{enumerate}
    \item Temporal Relation (``TR'')
    \begin{itemize}
        \item Ask about the sequence or simultaneity of actions.
        \item Distractors can involve: order swap, misplaced insertion/omission or wrong simultaneity.
    \end{itemize}
    \item Causal Relation (``CR'')
    \begin{itemize}
        \item Given an action X and ask about its effect, or given an effect Y and ask about its cause.
    \end{itemize}
    \item Counterfactual Relationship (``CRS'')
    \begin{itemize}
        \item The question stem deliberately describes an incorrect event order, and the question asks for the result of this wrong order.
        \item The correct option must explicitly identify the error in the stem’s described order, provide the correct sequence of events, and state the true consequence/result under this correct sequence.
        \item For distractors, two continue the incorrect sequence and describe plausible but false results under that assumption. Two also recognize the error but describe incorrect corrections or outcomes inconsistent with the caption.
    \end{itemize}
\end{enumerate}
\\
\hline
\end{tabular}
\caption{Prompt template for Temporal-Causal Relation.}
\label{tab:tier2_prompt}
\end{table*}

\begin{table*}[ht]
\centering
\begin{tabular}{|p{0.97\linewidth}|}
\hline
\textbf{Prompt Template: Generating QA Pairs for Intentional Understanding} \\
\hline
You are an expert at designing multiple-choice questions (MCQs) about intention reasoning for fine-grained actions in videos.

\vspace{3mm}
\textbf{[Objective]}

Given the detailed caption annotation for the entire video, design \{N\_QUESTIONS\} MCQs that evaluate the understanding of the intentions behind actions.

\vspace{3mm}
\textbf{[Design Rules]}
\begin{enumerate}
    \item The question stem must address the main actor in the second person (``you'').
    \item The options must describe the main actor’s behavior in the first person (``I'', ``my'').
    \item Each question has 5 options, with exactly 1 correct; do not prefix options with A/B/C.
    \item The correct option is a minimal, faithful restatement of the caption; try not to invent or modify the intention details that are not supported by the caption.
    \item The distractors must be logically reasonable but not consistent with the video content.
    \item Do not mention ``caption'', ``annotation'', or ``text''. Use phrases like ``in my view'' or ``I see''.
    \item Different questions should cover different parts or phases of the video as much as possible, avoiding repetitive questioning.
    \item Questions must require temporal understanding of the video: they should not be answerable from a single static frame or from the question wording alone.
\end{enumerate}

\vspace{3mm}
\textbf{[Question Types]}
\begin{enumerate}
    \item Functional Intention (``FI'')
    \begin{itemize}
        \item Action-level: Ask about why a specific action is performed (its purpose or intended result).
        \item Atomic-level: Ask about why a specific action is performed in a particular way, focusing on one or more execution dimensions (agent/effector, interaction object \& tool, object part, manipulation type, dynamic qualities, action direction, object location, repetition count, etc.).
    \end{itemize}
    \item Global Intention (``GI'')
    \begin{itemize}
        \item Focus: What is the overall purpose or goal of the protagonist’s sequence of actions?
    \end{itemize}
    \item Counterfactual Intention (``CIT'')
    \begin{itemize}
        \item Focus: If the protagonist had a different intention, how should the action execution change?
    \end{itemize}
\end{enumerate}
\\
\hline
\end{tabular}
\caption{Prompt template for Intentional Understanding.}
\label{tab:tier3_prompt}
\end{table*}

\begin{table*}[ht]
\centering
\begin{tabular}{|p{0.97\linewidth}|}
\hline
\textbf{Prompt Template: Generating QA Pairs for Evaluative Judgment} \\
\hline
You are an expert at designing multiple-choice questions (MCQs) about evaluating the quality of execution for fine-grained actions in videos.

\vspace{3mm}
\textbf{[Objective]}

Given the detailed caption annotation for the entire video, design \{N\_QUESTIONS\} MCQs that evaluate the quality and effectiveness of action execution-covering agent/effector, interaction object and tool, object part, manipulation type, dynamic qualities, action direction, object location, and repetition count.

\vspace{3mm}
\textbf{[Design Rules]}
\begin{enumerate}
    \item The question stem must address the main actor in the second person (``you'').
    \item The options must describe the main actor’s behavior in the first person (``I'', ``my'').
    \item Each question has 5 options, with exactly 1 correct; do not prefix options with A/B/C.
    \item The correct option is a minimal, faithful restatement of the caption; try not to invent or modify the details that are not supported by the caption.
    \item The distractors must be logically reasonable but not consistent with the video content.
    \item Do not mention ``caption'', ``annotation'', or ``text''. Use phrases like ``in my view'' or ``I see''.
    \item Different questions should cover different parts or phases of the video as much as possible, avoiding repetitive questioning.
    \item Questions must require temporal understanding of the video: they should not be answerable from a single static frame or from the question wording alone.
\end{enumerate}

\vspace{3mm}
\textbf{[Question Types]}
\begin{enumerate}
    \item Process Monitoring (``PM'')
    \begin{itemize}
        \item Evaluate the stage, smoothness, or blockage of task execution in relation to the global intention.
        \item Identify whether progress is smooth, temporarily obstructed, failed, completed, or interrupted.
        \item Questions should focus on temporal stages and transitions rather than static conditions.
    \end{itemize}
    \item Strategy Evaluation (``SE'')
    \begin{itemize}
        \item Assess the appropriateness, efficiency, and coordination complexity of the chosen execution dimensions.
        \item Questions should analyze how well the chosen method fits the physical constraints, and how effectively it balances coordination and efficiency.
    \end{itemize}
    \item Counterfactual Evaluation (``CEU'')
    \begin{itemize}
        \item Explore how changing a specific execution detail (e.g., tool, force, direction, timing, or repetition) would alter the quality, efficiency, or success of the action.
    \end{itemize}
\end{enumerate}
\\
\hline
\end{tabular}
\caption{Prompt template for Evaluative Judgment.}
\label{tab:tier4_prompt}
\end{table*}
\begin{table*}[ht]
\begin{tabular}{|p{0.97\linewidth}|}
\hline
\textbf{Prompt Template for Open-Ended Evaluation (1/4)} \\
\hline
You are an expert at evaluating answers to open-ended questions about fine-grained video understanding and reasoning.

You will evaluate the model's response to the given question based on the complete fine-grained description of the video (``VideoCaption'') and the human-written correct answer (``CorrectAnswer''), and score the model's performance on two dimensions: ``correctness'' and ``detailedness'', each within the range of 0 to 10.

Use the VideoCaption as the primary factual reference. The CorrectAnswer is only an auxiliary reference; if VideoCaption and CorrectAnswer conflict, always follow the VideoCaption.

\vspace{2mm}
The open-ended task types include:
\begin{enumerate}
    \item Counterfactual Interaction (``CIA''): The question asks how an action that did not occur would be executed. Check whether the answer points out the error in the question and explains the correct way of execution.
    \item Counterfactual Relationship (``CRS''): The question describes an incorrect temporal or causal relationship. Check whether the answer identifies the error and provides the correct sequence and outcome.
    \item Functional Intention (``FI''): Focus on the intention behind a specific action.
    \item Global Intention (``GI''): Focus on the overall purpose or goal of the protagonist’s sequence of actions.
    \item Counterfactual Intention (``CIT''): Focus on how execution should change if the protagonist had a different intention.
    \item Process Monitoring (``PM''): Focus on the stage, smoothness, or blockage of task execution in relation to the global intention.
    \item Strategy Evaluation (``SE''): Focus on the appropriateness, efficiency, and coordination complexity of the chosen execution strategy.
    \item Counterfactual Evaluation (``CEU''): Focus on how changing a specific execution detail would alter the quality, efficiency, or success of the action.
\end{enumerate}
\\
\vspace{2mm}
\textbf{Phase 1: Zero-Tolerance Check (for CIA / CRS)}

If the TaskType is CIA or CRS, the CandidateAnswer must explicitly reject OR implicitly correct the false premise in the question stem.
\begin{itemize}
    \item If the model explicitly identifies that the question’s description is wrong or inconsistent with the video, OR the model implicitly corrects the premise by accurately describing the true events in the video that contradict the false assumption, proceed to Phase 2.
    \item If the model passively accepts the false premise as true and hallucinates based on it (i.e., acquiescence), assign Correctness=0 and Detailedness=0, and skip directly to the final output.
\end{itemize}

\vspace{2mm}
If the TaskType is not CIA or CRS, skip Phase 1 and proceed directly to Phase 2.
\\
\hline
\end{tabular}
\caption{Prompt template for GPT-assisted evaluation of CFG-Bench (Part 1).}
\label{tab:gpt_prompt_1}
\end{table*}

\begin{table*}[ht]
\centering
\begin{tabular}{|p{0.97\linewidth}|}
\hline
\textbf{Prompt Template for Open-Ended Evaluation (2/4)} \\
\hline
\textbf{Phase 2: Scoring on Two Dimensions}\\
\textbf{Dimension 1: Correctness (0–10 points)}\\
Evaluate whether the CandidateAnswer is factually consistent with the VideoCaption and truly answers the reasoning demand of the Question.

\vspace{2mm}
\textbf{Correctness Rating Criteria:}\\
\textbf{9–10 points (Perfect)} \\
- Completely accurate. Matches the VideoCaption facts and the core logic of the CorrectAnswer.\\
- For counterfactual questions, accurately corrects the factual error in the premise (either explicitly through refutation or implicitly through factual description) and provides the true execution details based on the video.\\
- No critical errors about who did what, when, where, why, or with what.

\vspace{2mm}
\textbf{7–8 points (High)}\\
- Correct main conclusion and key reasoning steps.\\
- May have minor omissions or small inconsistencies compared to the CorrectAnswer, but no clear contradictions with the VideoCaption.\\
- Correctly addresses the intended reasoning type (e.g., intention, process, strategy, counterfactual).

\vspace{2mm}
\textbf{5–6 points (Medium)}\\
- Partially correct: captures the general idea or part of the reasoning chain, but misses important steps or conditions.\\
- May include one notable mistake or several minor slips, or mix correct and incorrect causal/temporal links.\\
- Still shows some meaningful alignment with the Question and VideoCaption.

\vspace{2mm}
\textbf{3–4 points (Low)}\\
- Significant errors: misinterprets the core of the Question or misreads key facts from the VideoCaption.\\
- Multiple incorrect claims about actions, actors, intentions, or temporal/causal relations.\\
- Reasoning type alignment is weak: the answer often talks about the wrong aspect (e.g., static appearance instead of intention).

\vspace{2mm}
\textbf{1–2 points (Poor)}\\
- Mostly irrelevant to the Question or heavily hallucinates facts not supported by the VideoCaption.\\
- Offers almost no correct reasoning about the video.

\vspace{2mm}
\textbf{0 points (Fail)}\\
- Fails the Phase 1 Check for CIA/CRS (does not explicitly identify the false premise), or the response is pure gibberish and cannot be meaningfully evaluated.
\\
\hline
\end{tabular}
\caption{Prompt template for GPT-assisted evaluation of CFG-Bench (Part 2).}
\label{tab:gpt_prompt_2}
\end{table*}

\begin{table*}[ht]
\centering
\begin{tabular}{|p{0.97\linewidth}|}
\hline
\textbf{Prompt Template for Open-Ended Evaluation (3/4)} \\
\hline
\textbf{Dimension 2: Detailedness (0–10 points)}\\
Evaluate whether the CandidateAnswer is comprehensive, precise, and focused when addressing the required reasoning type, including conditions, nuances, and evidence from the VideoCaption.

\vspace{2mm}
\textbf{Detailedness Rating Criteria:}\\
\textbf{9–10 points (Rich)}\\
- Thoroughly explains the relevant ``why” and/or ``how” for the given reasoning type (intention, process, strategy, counterfactual, etc.).\\
- Uses multiple specific details from the VideoCaption (objects, movements, sequence of steps, outcomes, constraints).\\
- Comparable in depth and nuance to the CorrectAnswer.

\vspace{2mm}
\textbf{7–8 points (Good)}\\
- Covers the main reasoning steps and most key conditions.\\
- Includes several concrete details from the VideoCaption, though may miss some secondary nuances.\\
- Explanation is coherent and informative, but not fully exhaustive.

\vspace{2mm}
\textbf{5–6 points (Average)}\\
- Provides a generally correct explanation but remains high-level or somewhat generic.\\
- Mentions some relevant elements (e.g., main intention or main cause–effect link) but lacks fine-grained detail or specific evidence.\\
- Limited use of explicit cues from the VideoCaption.

\vspace{2mm}
\textbf{3–4 points (Weak)}\\
- Brief or superficial; focuses on only one part of the reasoning while omitting other important aspects.\\
- Contains very few concrete details from the VideoCaption.\\
- Does not fully engage with the required reasoning type (e.g., just states the result without explaining why).

\vspace{2mm}
\textbf{1–2 points (Poor)}\\
- Extremely short, vague, or empty.\\
- Almost no usable reasoning or concrete detail.

\vspace{2mm}
\textbf{0 points (Fail)}\\
- Fails the Phase 1 Check for CIA/CRS (thus Detailedness must also be 0), or is pure gibberish with no coherent content.
\\
\hline
\end{tabular}
\caption{Prompt template for GPT-assisted evaluation of CFG-Bench (Part 3).}
\label{tab:gpt_prompt_3}
\end{table*}

\begin{table*}[ht]
\centering
\begin{tabular}{|p{0.97\linewidth}|}
\hline
\textbf{Prompt Template for Open-Ended Evaluation (4/4)} \\
\hline
\textbf{Additional Policies}\\
- Expression tolerance: Paraphrasing or surface wording differences are acceptable if the meaning is equivalent.  \\
- Multiple valid answers: Different correct phrasings expressing the same reasoning are equally valid.  \\
- No reward for verbosity: Long but off-topic or repetitive text does not increase the Detailedness score. Only relevant, grounded details count. \\ 
- Correctness first: If an answer is seriously incorrect, it should not receive high scores even if it is long or detailed.  \\
- Omissions vs. errors: Omitting some content is less severe than stating it incorrectly. Explicit errors should penalize Correctness more than simple omissions.
\\
\vspace{2mm}
\textbf{Uncertain or Incomplete Evidence Policy}

Even if the VideoCaption does not explicitly state the reasoning or outcome, you must infer the most plausible interpretation from the available evidence and context.\\
- Always provide a reasoned judgment rather than defaulting to pure uncertainty.\\
- Use lower scores to reflect weak or incomplete evidence instead of skipping evaluation.\\
- Avoid outputs like ``Underspecified''; always justify your stance with textual evidence or reasonable inference.

\\
\vspace{2mm}
\textbf{Context}\\
- VideoCaption: \{caption\}\\
- Question: \{question\}\\
- CorrectAnswer: \{correct\_answer\}\\
- TaskType:\{task\_type\}\\
- CandidateAnswer: \{result\}

\\
\vspace{2mm}
\textbf{Output (STRICT JSON)}\\
Return only:\\
\{
\\
  ``evidence\_spans'': ``short quotes from VideoCaption that support your judgment'',\\
  ``correctness\_reasoning'': ``explanation of your correctness score'',\\
  ``detailedness\_reasoning'': ``explanation of your detailedness score'',\\
  ``correctness'': [integer 0-10],\\
  ``detailedness'': [integer 0-10],\\
\}
\\
\hline
\end{tabular}
\caption{Prompt template for GPT-assisted evaluation of CFG-Bench (Part 4).}
\label{tab:gpt_prompt_4}
\end{table*}
\end{document}